\documentclass[11pt, a4paper, onecolumn, copyright, goog]{google}

\usepackage[authoryear, sort&compress, round]{natbib}
\usepackage{multirow}
\usepackage{amsmath}
\usepackage{amsthm}
\usepackage[ruled,vlined]{algorithm2e}
\newtheorem{proposition}{Proposition}
\usepackage{pgfplots}
\pgfplotsset{compat=1.18}
\usepackage{tikz}
\usetikzlibrary{patterns, fit, positioning, shapes.geometric, arrows.meta}
\usepackage{subcaption}
\usepackage{wrapfig}

\newcount\Comments  %
\newif\ifshowcomments
\showcommentstrue
\usepackage{color}
\definecolor{darkred}{rgb}{0.9,0,0}
\definecolor{darkgreen}{rgb}{0,0.5,0}
\definecolor{darkblue}{rgb}{0,0,0.7}
\definecolor{purple}{rgb}{.6, 0,.6}
\definecolor{orange}{rgb}{1.0,0.64,0}
\newcommand{\kibitz}[2]{\ifshowcomments\ifnum\Comments=1\textcolor{#1}{#2}\fi\fi}

\uselogo{} 

\title{Give the Drafter a Second Chance to Guess:
Parallel Reconsideration for Speculative Decoding}

\correspondingauthor{el72@rice.edu}

\reportnumber{} 

\author[1,2,*]{Hoang Anh Duy Le}
\author[1]{Huibing Dong}
\author[1]{Alireza Khadem}
\author[3]{Ziteng Sun}
\author[1]{Xiang Song}
\author[4]{Yizheng Huang}
\author[1]{Soroush Radpour}
\author[2]{Anshumali Shrivastava}
\author[1]{Ye He}

\affil[1]{\thepa{}{}}
\affil[2]{Rice University}
\affil[3]{Google Research}
\affil[4]{Google DeepMind}
\affil[*]{Work done at Google.}

\begin{abstract}
    Speculative decoding accelerates large language model inference with a small
    drafter, whose per-guess acceptance rate determines the achievable speedup.
    Feature-level drafters share a design choice: every token, easy or hard, is
    drafted under a \emph{single} forward pass. Many of the tokens the target rejects
    are ones the drafter is \emph{itself unsure about} yet the target
    answers decisively. To
    address this, we introduce \textbf{Parallel Reconsideration}: guided by its own uncertainty, the drafter \emph{reconsiders} its
    hesitant guesses with extra passes that run \emph{in parallel} with the
    one-pass draft, adding negligible latency. A co-designed training recipe and serving-time drafting
    algorithm append these reconsidered guesses to the original draft tree as refined
    candidates.
    Across models and benchmarks, on average, Parallel Reconsideration improves over the
    baselines by \textbf{5.3\%} to
    \textbf{10.2\%} on acceptance length and by \textbf{3.5\%} to \textbf{9.2\%}
    on speedup.
\end{abstract}

\begin{document}

\maketitle

\section{Introduction}
\label{sec:intro}

Large language models (LLMs) have become a pivotal component of modern
computing, powering assistants, search, and the fast-growing class of
\emph{agentic} systems that plan, call tools, and write code over long
horizons~\citep{anthropic_claudecode, openai_codex, google_antigravity, moonshot_kimik3}.
Yet this shift exposes a stubborn cost: in
the agentic era even a simple coding task can unfold into
generated tokens across many chained model calls, and each of those tokens is produced
\emph{one at a time}. This is not an implementation artifact but the nature of
\textbf{autoregressive decoding}: every token is conditioned on all previous
tokens, so a single forward pass of a multi-billion-parameter model yields
exactly one token, and end-to-end latency scales with the length of the
generation~\citep{pope2023efficiently}. As outputs grow longer,
this sequential bottleneck, rather than raw compute, increasingly governs how
fast a model can respond.

\textbf{Speculative decoding} directly mitigates this bottleneck~\citep{leviathan2023fast, xia2023speculative}.
A cheap \emph{drafter} proposes several candidate tokens, and the target model
\emph{verifies} them in parallel in a \emph{single} forward pass. A \emph{modified rejection sampling} step keeps the longest run of
drafted tokens the target agrees with, provably preserving its output
distribution exactly~\citep{leviathan2023fast}; the saving comes from
\emph{parallelism}, as
$k$ accepted tokens cost one target pass instead of $k$ expensive ones. The drafter thus decides
everything: the more tokens it gets right per round, the greater the speedup. One notable class is the EAGLE
family~\citep{li2024eagle, li2024eagle2, li2025eagle3}, which trains a lightweight
drafter that predicts at the \emph{feature} level rather than the token level,
reusing the target's hidden states to draft with a single small decoder layer and
setting the standard for modern speculative decoding.

This efficiency, however, comes with a design choice shared by feature-level drafters:
\textbf{every token, easy or hard, is drafted with the same fixed budget of a single
    forward pass through a small decoder.}\footnote{For ease of discussion we
    assume a single-layer, EAGLE-style drafter throughout; the ideas extend directly
    to deeper or otherwise structured drafters, where the fixed budget is one pass
    through the whole drafter.} Some proposed tokens the target
rejects are not ones the draft gets confidently wrong; they are
tokens the draft is \emph{itself unsure about} even though the target has a clear answer
there. We hypothesize that part of this loss stems from how much the draft can
\emph{compute}: a single pass of one small layer may not suffice to
settle these hard tokens, yet the draft must commit and move on regardless.

The obvious fix is to give the drafter more capacity, e.g. more layers.
Yet a bigger drafter slows the entire
round and eats into the speedup that is the whole point of speculative decoding.
This exposes the real tension:
\textbf{how can the drafter spend more compute on the tokens that need it,
    without paying for that compute on the tokens that do not?} A viable answer must
be \emph{selective}, spending the extra compute only at the uncertain positions; \emph{(near-)free}, adding negligible wall-clock latency
over the ordinary draft; and \emph{lossless}, leaving the verification untouched.

We answer this with \textbf{Parallel Reconsideration}, which grants the drafter a
\emph{second chance to guess} on the tokens it is unsure about. The drafter's own
uncertainty is the signal: where it is already decisive, the guess is committed
as usual; where it hesitates, the drafter is given more attempts
to revise its guess into a better one before committing. Moreover, this reconsideration
is nearly \emph{free}: the additional attempts are batched into the same forward
pass as the ordinary draft, adding only negligible parallel overhead at inference time.
Our contributions are as follows:
\begin{itemize}[leftmargin=*, noitemsep, topsep=0pt]
    \item \textbf{Parallel Reconsideration.}
          We introduce uncertainty-guided \emph{draft-time scaling} to feature-level
          drafters: instead of a fixed one-pass-per-token budget, the drafter scales
          its computation per position according to its own uncertainty. Parallel Reconsideration reconsiders the uncertain guesses with additional passes that run
          \emph{in parallel} with the ordinary draft, so the added computation is nearly
          free (Section~\ref{sec:method}).
    \item \textbf{A co-design of training and serving.} We
          design a training recipe under which the reconsideration branch deliberately learns to
          correct the drafter's mistakes without degrading its ordinary drafting
          branch (Section~\ref{sec:training}). In serving, we design a drafting algorithm that grafts the
          reconsidered guesses onto the ordinary draft tree as \emph{additional}
          candidates (Section~\ref{sec:serving}).
    \item \textbf{Extensive empirical evaluation.} Across models and
          benchmarks, Parallel Reconsideration improves acceptance length over the
          baselines by \textbf{5.3--10.2\%} on average across models, draft
          budgets, and sampling temperatures, at a nearly identical per-round drafting
          cost. We further show that spending the same compute
          on a wider draft or a second decoder layer does not match it, and ablate each
          component of the design
          (Section~\ref{sec:experiments}).
\end{itemize}

\section{Background and Motivation}
\label{sec:motivation}


\subsection{Preliminaries and the cost model}
\label{sec:moti_prelim}

We consider an EAGLE-style drafter~\citep{li2024eagle} that predicts at the
feature level. Let $f_p^{0}$ denote the draft feature at speculative position
$p$ produced by a single forward pass of the drafter's single decoder layer $M$, let
$e(t)$ be the embedding of token $t$, let $\mathrm{fc}(g, e)$ be EAGLE's fusion of a
feature $g$ with a token embedding $e$ (concatenate, then project), and let
$\mathrm{head}(\cdot)$ be the LM head. To advance one position, the drafter samples a
token $t_{p+1} \sim \mathrm{head}(f_p^{0})$ from the current feature and predicts the
next feature by
$f_{p+1}^{0} = M(\mathrm{fc}(f_p^{0}, e(t_{p+1})))$, one \emph{draft step}.
Each round, the drafter generates candidate tokens in $\gamma$ steps sequentially, incurring a
drafting cost
\begin{equation}
    T_{\text{draft}} = \gamma \cdot t_{\text{step}},
    \label{eq:tdraft}
\end{equation}
where $t_{\text{step}}$ is the latency of a single forward pass of $M$;
the drafting cost therefore grows linearly with $\gamma$.
The target model then verifies the candidates in one forward pass,
costing $T_{\text{verify}}$, and accepts the longest prefix it agrees with. The
expected number of tokens accepted per round is the \emph{acceptance length}
$\tau$, and the amortized cost per accepted token is
\begin{equation}
    \frac{T_{\text{draft}} + T_{\text{verify}}}{\tau}
    = \frac{\gamma\, t_{\text{step}} + T_{\text{verify}}}{\tau}.
    \label{eq:costmodel}
\end{equation}
The goal is therefore to \textbf{raise the acceptance length $\tau$ while
    holding $T_{\text{draft}} + T_{\text{verify}}$ on par with prior methods}: a
drafter that raises $\tau$ by inflating $\gamma\, t_{\text{step}}$ can easily
lose more in the numerator than it gains in the denominator. Underlying this
formulation is an implicit assumption that prior work has left unquestioned:
each step runs $M$ \emph{exactly once}, so every token, easy or hard,
is drafted from a single forward pass with no opportunity to revise.

\subsection{Where does the drafter lose acceptance?}
\label{sec:moti_diagnosis}

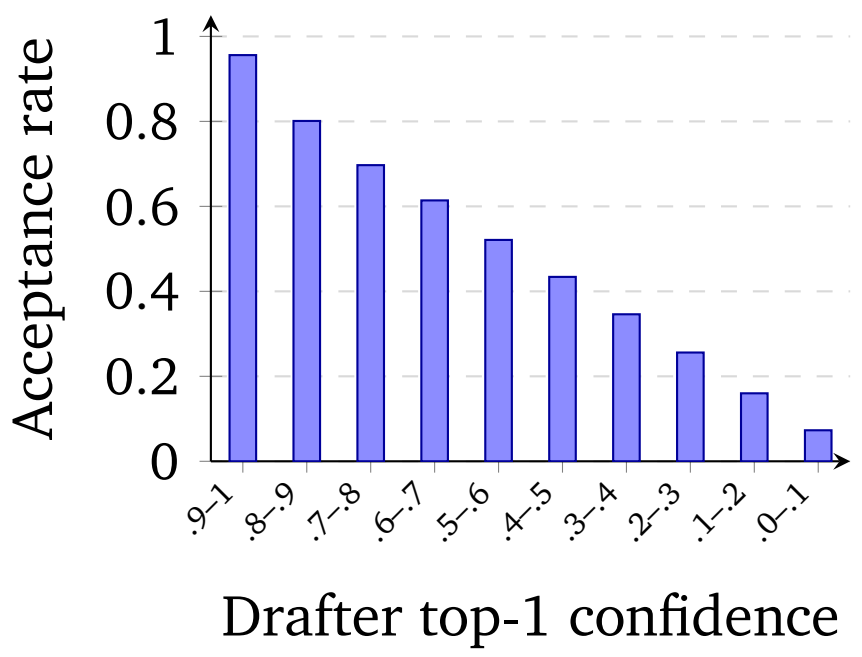
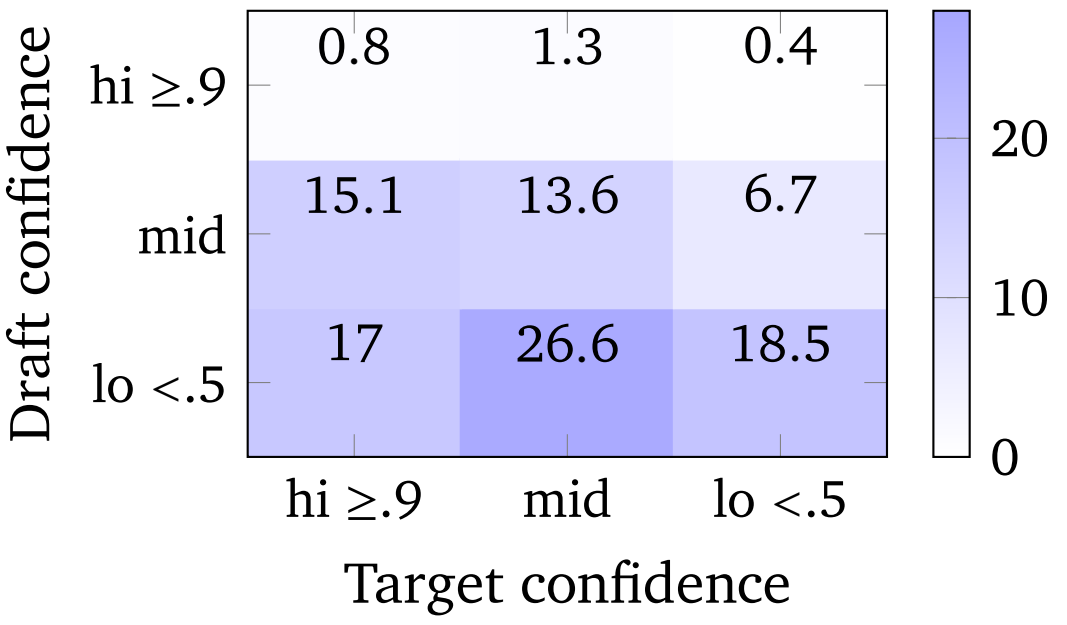
\begin{figure*}[t]
    \centering
    \hfill
    \begin{subfigure}[b]{0.35\textwidth}
        \centering
        \begin{tikzpicture}
            \begin{axis}[
                    width=\linewidth,
                    height=0.78\linewidth,
                    ybar,
                    bar width=5pt,
                    ymin=0, ymax=1.05,
                    xmin=0.5, xmax=10.5,
                    xtick={1,2,3,4,5,6,7,8,9,10},
                    xticklabels={{.9--1},{.8--.9},{.7--.8},{.6--.7},{.5--.6},{.4--.5},{.3--.4},{.2--.3},{.1--.2},{.0--.1}},
                    x tick label style={rotate=45, anchor=east, font=\tiny},
                    ytick={0,0.2,0.4,0.6,0.8,1.0},
                    yticklabel style={font=\small},
                    ylabel={Acceptance rate},
                    xlabel={Drafter top-$1$ confidence},
                    ylabel style={font=\normalsize},
                    xlabel style={font=\normalsize},
                    enlarge x limits=0.03,
                    axis lines=left,
                    ymajorgrids=true,
                    grid style={dashed, gray!30},
                ]
                \addplot[draw=blue!60!black, fill=blue!45] coordinates {
                        (1,0.956) (2,0.801) (3,0.697) (4,0.614) (5,0.521)
                        (6,0.434) (7,0.346) (8,0.256) (9,0.160) (10,0.073)
                    };
            \end{axis}
        \end{tikzpicture}
        \caption{Acceptance vs.\ drafter confidence.}
        \label{fig:accept_vs_conf}
    \end{subfigure}
    \hfill
    \begin{subfigure}[b]{0.35\textwidth}
        \centering
        \begin{tikzpicture}
            \begin{axis}[
                    width=\linewidth,
                    height=0.78\linewidth,
                    enlargelimits=false,
                    colormap={blues}{color(0)=(white) color(1)=(blue!35)},
                    colorbar,
                    colorbar style={font=\small, width=0.24cm},
                    point meta min=0, point meta max=28,
                    xtick={0,1,2},
                    xticklabels={hi $\ge$.9, mid, lo $<$.5},
                    ytick={0,1,2},
                    yticklabels={lo $<$.5, mid, hi $\ge$.9},
                    x tick label style={font=\small},
                    y tick label style={font=\small},
                    xlabel={Target confidence},
                    ylabel={Draft confidence},
                    xlabel style={font=\normalsize},
                    ylabel style={font=\normalsize},
                    axis on top,
                    nodes near coords style={font=\small\bfseries, text=black},
                ]
                \addplot[
                    matrix plot*,
                    mesh/cols=3,
                    point meta=explicit,
                    nodes near coords={\pgfmathprintnumber\pgfplotspointmeta},
                ] coordinates {
                        (0,0) [17.0] (1,0) [26.6] (2,0) [18.5]
                        (0,1) [15.1] (1,1) [13.6] (2,1) [6.7]
                        (0,2) [0.8]  (1,2) [1.3]  (2,2) [0.4]
                    };
            \end{axis}
        \end{tikzpicture}
        \caption{Draft vs.\ target confidence (\% of rejects).}
        \label{fig:draft_x_target}
    \end{subfigure}
    \hfill\null
    \caption{\textbf{Two observations on where an EAGLE-style drafter loses
            acceptance.}
        \textbf{(\subref{fig:accept_vs_conf})} Acceptance rate versus the drafter's own
        top-$1$ confidence, which falls from $95.6\%$ on confident positions to $7.3\%$ on
        the least confident.
        \textbf{(\subref{fig:draft_x_target})} Drafter confidence against target confidence
        over the rejected tokens (\% of the reject pool): the drafter
        is unsure on $62.1\%$ of them, and $17.0\%$ of the pool is unsure at a
        decisive target ($c^{\mathrm{t}}_p \ge 0.9$).}
    \label{fig:diagnosis}
\end{figure*}

To understand where acceptance is lost, we analyze a trained EAGLE-3.1 drafter for
Llama-3.1-8B-Instruct on a held-out split.
For each drafted position $p$, we read off two quantities: the \emph{drafter
    confidence} $c^{\mathrm{d}}_p = \max_t \mathrm{head}(f_p^{0})_t$, its own top-$1$
probability, and the \emph{target confidence}
$c^{\mathrm{t}}_p = \max_t \pi_{\mathrm{tgt}}(t \mid x_{<p})$, the top-$1$
probability of the target's next-token distribution at the same position. The
drafter confidence measures how sure the drafter is; the target confidence
measures how \emph{resolvable} the position is, i.e. when $c^{\mathrm{t}}_p$ is
high, a single correct token exists and can, in principle, be predicted by the drafter.
Figure~\ref{fig:diagnosis} summarizes two observations relating these
quantities to acceptance.

\paragraph{Observation 1: confidence measures the opportunity cost of a drafted token.}
Acceptance tracks $c^{\mathrm{d}}_p$ closely
(Figure~\ref{fig:accept_vs_conf}): positions the drafter is confident about
($c^{\mathrm{d}}_p \ge 0.9$) are accepted $95.6\%$ of the time, and acceptance
falls steadily as its confidence drops: to $0.52$ at $0.5$--$0.6$ and $0.07$
in the lowest bucket. The drafter is thus seldom confidently wrong; where it loses
acceptance, it was already unsure. The drafter's own confidence, available at draft time,
therefore indicates how likely the target is to reject the token it drafts.

\paragraph{Observation 2: a fair share of rejections is recoverable, but confidence does not say which.}
Among rejected tokens
(Figure~\ref{fig:draft_x_target}), the drafter is unsure on $62.1\%$ of them, and
$43.6\%$ fall at a target that still favors one token ($c^{\mathrm{t}}_p \ge 0.5$),
$17.0\%$ of rejected tokens at a decisive one ($c^{\mathrm{t}}_p \ge 0.9$), against a comparable
$18.5\%$ where the target is unsure ($c^{\mathrm{t}}_p < 0.5$). A fair share of the
acceptance the drafter loses is thus recoverable: at those positions a token the target
would have accepted was there to be drafted. But the drafter cannot tell which: its
confidence is equally low at a decisive target and at an ambiguous one.

\subsection{Spend compute where the drafter is unsure}
\label{sec:moti_insight}

The two observations compose into a single reading: the drafter loses acceptance on tokens
it is itself unsure about (Observation~1), on a fair share of which the target nonetheless
has a decisive answer (Observation~2). It is not confidently wrong at these positions, and
a correct token is there to be drafted, yet it is \textbf{committed prematurely}: a single draft
step is not enough to resolve the uncertain position $p$ before the drafter must commit
its guess and move to the next position $p+1$. What the drafter is missing, then, is not a
better objective but more computation:

\begin{quote}
    \emph{Give the drafter more compute where it is unsure, and none extra where
        it is not.}
\end{quote}

\noindent Within feature-level drafters, prior work mostly refines \emph{what} the
drafter computes in a single step rather than \emph{how much} it computes, which stays
fixed at one forward pass of $M$ per position.
A naive bigger drafter with more compute is self-defeating under
Eq.~\eqref{eq:costmodel}, since it raises $t_{\text{step}}$ on \emph{every}
position, including the confident majority that one step already resolves,
inflating $T_{\text{draft}}$ faster than it lifts $\tau$
(see Section~\ref{sec:exp_design} for details). We therefore scale draft-time compute
\emph{selectively}, spending extra draft steps only on the uncertain positions where the
acceptance lost is recoverable, and leaving the confident majority at one forward pass, as
\textbf{Parallel Reconsideration} (PR). Section~\ref{sec:experiments}
revisits these observations and analyzes whether Parallel Reconsideration indeed
helps on the raised challenges.

\section{Parallel Reconsideration}
\label{sec:method}

We introduce \textbf{Parallel Reconsideration} with the reconsideration step
(Section~\ref{sec:method_reconsider}), the learned self-query that lets the drafter refine the single-pass
draft rather than growing the next one (Section~\ref{sec:method_resolve}), the
learned gate that decides which positions have a chance of drafting a better token through reconsideration (Section~\ref{sec:method_gate}), and the parallel execution that keeps it cheap
(Section~\ref{sec:method_parallel}). We describe the method with a single
reconsideration step for clarity, though in practice,
the drafter may take more than one.

\subsection{The reconsideration step}
\label{sec:method_reconsider}

Recall that the ordinary draft step at position $p$ reads the one-pass feature
$f_p^{0}$, samples the next token $t_{p+1} \sim \mathrm{head}(f_p^{0})$, and
advances to position $p+1$ via $f_{p+1}^{0} = M(\mathrm{fc}(f_p^{0},
    e(t_{p+1})))$. Given an uncertain feature $f_p^{0}$ and $Q$ a learned self-query (defined below), \emph{Reconsideration}
spends one additional forward pass to refine it into
$f_p^{1}$, from which the token is re-sampled:
\begin{equation}
    f_p^{1} = M\!\left(\mathrm{fc}(f_p^{0}, Q(f_p^{0}))\right),
    \qquad
    \tilde t_{p+1} \sim \mathrm{head}(f_p^{1}),
    \label{eq:reconsider}
\end{equation}

\subsection{The self-query}
\label{sec:method_resolve}

In an EAGLE-style drafter, current feature $f_p^{0}$ does not determine the next feature on its own, since
different sampled tokens $t_{p+1}$ continue into different
$f_{p+1}^{0}$~\citep{li2024eagle}, so the drafter fuses $e(t_{p+1})$ to tell $M$ which branch to advance
along.  Reconsideration, however, stays at $p$ without advancing to re-decide the very token
$t_{p+1}$ the drafter is unsure about.
We therefore give reconsideration a \emph{steering embedding}: a learned \emph{self-query} $Q(f_p^{0})$, with
$Q : \mathbb{R}^d \to \mathbb{R}^d$, encoding
what a second look at an uncertain state should attend to in order to refine $f_p^{0}$ instead of the naive one-pass draft of next token's feature.

\subsection{The reconsideration gate}
\label{sec:method_gate}

Not every uncertain position is worth another pass, and $c^{\mathrm{d}}_p$ does not say
which are: it reads the same where the target is decisive and where it is ambiguous
(Observation~2). Recoverability has to be learned from the feature itself. Writing
$f_p^{0},\dots,f_p^{R}$ for the features at each reconsideration level, up to at most
$R$, a small gate at level $r$ reads each and emits a halt probability:
\begin{equation}
    \lambda_r = \sigma\!\left(\mathrm{gate}(f_p^{r})\right),
    \qquad
    q(r) = \lambda_r \prod_{s<r}(1-\lambda_s).
    \label{eq:gate}
\end{equation}
Every level asks the same question, whether another pass still has a chance of drafting a
better token, so $q(r)$ is the chance a position reaches level $r$ and stops there.

\subsection{Reconsidering in parallel}
\label{sec:method_parallel}

A naive Reconsideration adds forward passes, and run naively in series each one stalls the draft at $p$
until $f_p^{1}$ is computed: a draft of $\gamma$ positions with $N$ reconsidered ones
costs $(\gamma+N)\,t_{\mathrm{step}}$ instead of $\gamma\,t_{\mathrm{step}}$
(Figure~\ref{fig:coforward}, top). However, the one-pass draft step and its reconsideration,
\begin{equation}
    f_{p+1}^{0} = M\big(\mathrm{fc}(f_p^{0}, e(t_{p+1}))\big),
    \qquad
    f_p^{1} = M\big(\mathrm{fc}(f_p^{0}, Q(f_p^{0}))\big),
    \label{eq:coforward}
\end{equation}
are disjoint, so the two
can be batched into a single forward of $M$: \emph{co-forwarding} rather than
serializing (Figure~\ref{fig:coforward}, bottom). At the small batch sizes of
latency-oriented decoding that forward is memory-bandwidth bound, so the added rows cost a negligible $\epsilon$ overhead (Appendix~\ref{app:method_parallel}).
Drafting thus becomes \emph{interleaved}: the ordinary steps grow the draft tree as
before, and in parallel, a reconsidered position adds its $\tilde t_{p+1}$ as a reconsidered child
beside the one-pass ones. Both kinds are draft candidates, pruned and verified together
under the standard speculative-decoding verification in serving time.
\section{Training Parallel Reconsideration}
\label{sec:training}

The single decoder layer $M$ must serve two heads at once: the one-pass
$\mathrm{head}(f_p^{0})$ that guesses and advances the draft to position $p+1$, and the
reconsidered $\mathrm{head}(f_p^{1})$ that revises a premature guess.
Neither can be trained merely as an intermediate step toward the other, because at
serving \emph{both} are read out and committed to the tree: the one-pass head guesses the
token at most positions and the reconsidered head revises it at the uncertain ones. Training must therefore keep
the one-pass head as accurate as an ordinary drafter while pushing the reconsidered head
to win on the recoverable tokens single pass misses, and doing so is delicate because the two
share the feature $f_p^{0}$. Supervise only the
reconsidered head and the one-pass head degrades, overloading all tokens to be reconsidered and losing the ordinary-drafter quality on the easy tokens;
supervise both to match the target with equal weight and gradient descent finds
the trivial solution $f_p^{1}\!\approx\!f_p^{0}$, so the reconsidered token merely echoes
the same one-pass token. Either failure would undercut
speculative-decoding quality (Appendix~\ref{app:train_ablation}). We therefore hold the
one-pass head to ordinary-drafter quality with an unconditional loss, and supervise each
reconsidered feature only at the positions and depths where it is actually used.

\textbf{One loss, every depth.} Thus during training, we predict token at every depth $r$ and score it with $\ell_p^{r} = \mathrm{CE}(\mathrm{head}(f_p^{r}), t_{p+1}^{\star})$, then weight
each depth's loss by the mass $q(r)$ the gate routes to it, while leaving the one-pass depth
unweighted. The gate has already judged the chance each depth is the one that settles the
token (Eq.~\eqref{eq:gate}), so $q(r)$ is the natural weight: a depth earns gradient in
proportion to how often it is the one asked to answer. Each depth's loss
also shapes where the gate halts, pulling its mass toward whichever depth reads out the
token best. Weighting alone, however, leaves the gate free to abandon a depth entirely (
a depth it stops routing to receives no gradient), so we add an
\emph{entropy} bonus that keeps every depth in use. With $\mathrm{sg}[\cdot]$ the stop-gradient, we have the training objective:
\begin{equation}
    \mathcal{L} = \sum_{p}
    \Big(
    \underbrace{\ell_p^{0}}_{\text{one-pass}}
    \;+\;
    \underbrace{\sum_{r\ge1} \mathrm{sg}[q(r)]\,\ell_p^{r}}_{\text{reconsideration heads}}
    \;+\;
    \underbrace{\sum_{r\ge0} q(r)\,\mathrm{sg}[\ell_p^{r}]}_{\text{gate}}
    \;-\;
    \underbrace{\beta\, H\!\big(\mathbb{E}_p[\,q\,]\big)}_{\text{balance}}
    \Big),
    \label{eq:fullloss}
\end{equation}

\paragraph{The mixed cache.}A reconsidered
child commits the key/value of its parent's halt level rather than of parent's one-pass feature, so
a position deeper in the draft attends over a \emph{mixture}: one-pass keys and values at
the confident ancestors that were never reconsidered, reconsidered ones at the uncertain
ancestors that were. To adapt to this serving phenomenon, each position $j$ we sample a
halt depth $d_j \sim q$ and commit the key/value that depth produces,
\begin{equation}
    \kappa_j = \mathrm{kv}\!\left(f_j^{\,d_j}\right), \qquad d_j \sim q,
    \label{eq:haltkv}
\end{equation}
so position $p$ attends over the prefix $\kappa_{<p}$, resampled afresh each step.

Appendix~\ref{app:training} gives the full training recipe, including the two on-policy
rollouts the two heads are trained on, the per-depth weighting, and the analysis of the
balance term.

\section{Serving a Reconsidered Tree}
\label{sec:serving}

\begin{figure*}[t]
    \centering
    \includegraphics[width=0.95\textwidth]{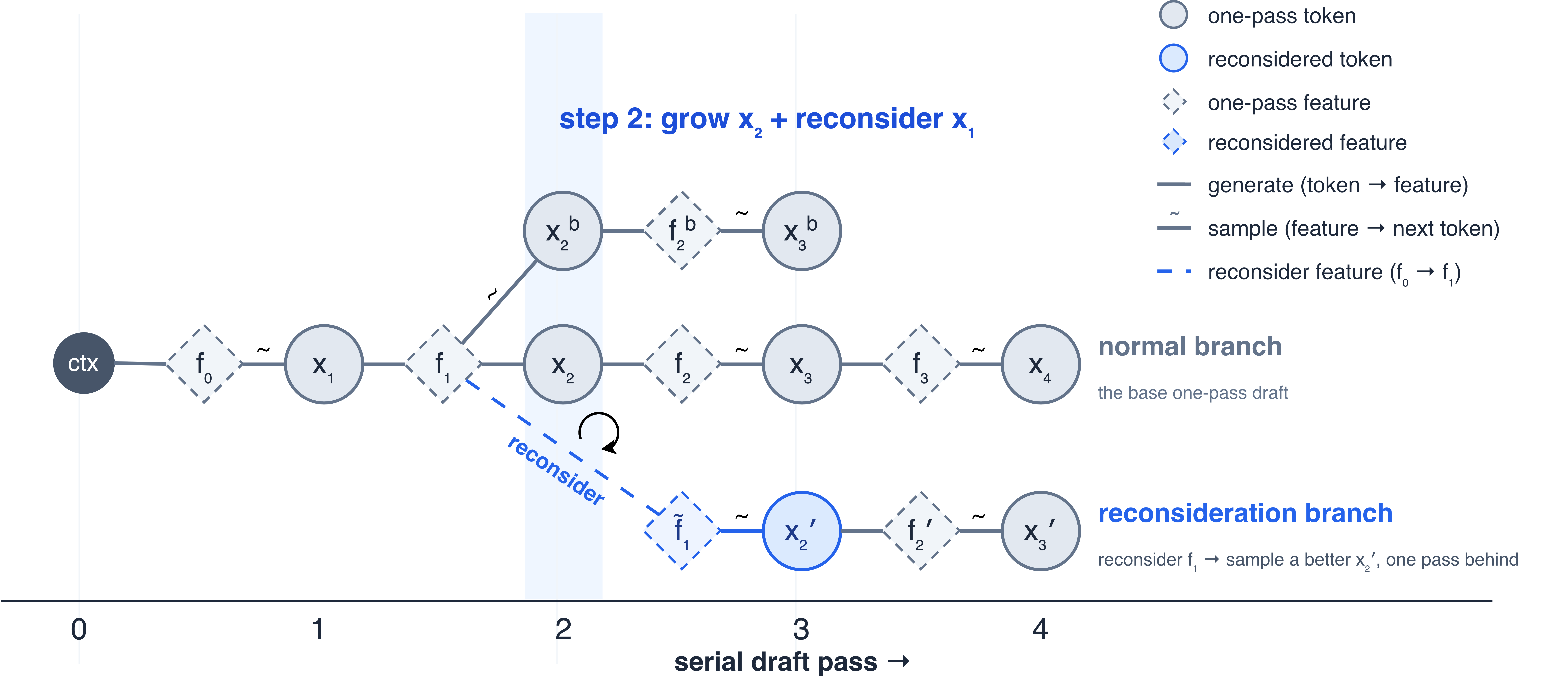}
    \caption{\textbf{Co-forwarding reconsideration on the draft tree.} Each token
        produces a feature (diamond) from which the next token is sampled (${\sim}$); the
        one-pass draft (gray, level $0$) forms the main branches, and tree width comes free
        because a node's top-$k$ children are all sampled from its single feature. The
        reconsideration pass (blue, level $1$) deepens a node's feature within the same
        batched forward as the next draft step so
        it proposes a corrected token (here $x_2'$) with no extra sequential draft step. The
        one-pass draft still reaches $x_4$ while the reconsidered branch grows in parallel.}
    \label{fig:servetree}
\end{figure*}

EAGLE-3.1 serves by building a \emph{dynamic draft tree}, grown over every round's
$\gamma$ draft steps. Each step forwards a batch of $B_w$ frontier nodes; every forwarded
node offers children sampled from its feature, and each node $v$ carries a score equal
to its cumulative path probability from the root,
$
    s(v) = \sum_{u \in \mathrm{path}(v)} \log \pi_{\mathrm{one}}(u \mid \mathrm{parent}(u)),
$
where $\pi_{\mathrm{one}}$ is the one-pass distribution of the node's parent. After
$\gamma$ steps, a single global cut keeps the $B$ highest-scoring nodes and prunes the
rest.

Parallel Reconsideration keeps that procedure and has a second one in parallel. Alongside the
frontier-growing nodes the drafter holds a \emph{reconsideration pool} $\mathcal{C}$ of nodes that have not yet halted. A draft step
now has two things to do, advance the frontier and reconsider the pool, and it does them
at once: the $B_w$ highest-scoring frontier nodes move the draft to the next position,
and the $m$ highest-bidding nodes of $\mathcal{C}$ are reconsidered, within the same forward batch of $M$
(Section~\ref{sec:method_parallel}), so the $\gamma$-step growth of Eq.~\eqref{eq:tdraft} is untouched.

Which $m$ nodes are reconsidered follows from what reconsideration changes: on the tree,
reconsidering a node $v$ leaves $v$ itself fixed and revises the distribution it offers
its children (Section~\ref{sec:method_reconsider}). A node is therefore worth
reconsidering when the draft is likely to reach it and its one-pass continuation is
likely to be wrong, which together are the probability that the draft breaks immediately
after $v$. We rank the pool by
\begin{equation}
    \mathrm{bid}(v) = \underbrace{s(v)}_{\log \Pr[\text{reach } v]}
    \;+\; \underbrace{\log\!\big(1 - \max\nolimits_t\, \pi_{\mathrm{one}}(t \mid v)\big)}_{\log \Pr[v\text{'s one-pass child is wrong}]}
    \;=\; \log \Pr[\text{the draft breaks at } v\text{'s children}],
    \label{eq:bid}
\end{equation}
and reconsider the top $m$. We note that the gate and the bid
are two complementary mechanisms: The gate asks whether another pass has a \emph{chance} of
drafting a better token, which confidence cannot answer (Observation~2); a node it declines to halt, meaning it has a chance of drafting a better child. The bid then asks, among the nodes
that have that chance, which are the most valuable to reconsider this step, and
confidence is what answers it: the less confident $v$'s one-pass guess, the likelier the
first draft guesses it wrong, and the more there is to gain by reconsidering it
(Observation~1).

A reconsidered node re-draws its children from the reconsidered feature, and
the token $\tilde t_v \sim \mathrm{head}(f_v^{d_v})$ it now proposes enters the tree as
an \emph{additional} child beside the ones its one-pass feature offered. Once node $v$ halts at $d_v$ with $\pi_{d_v}=\mathrm{head}(f_v^{d_v})$,
the draft has a new unified scoring over its original vs. reconsidered children,
\begin{equation}
    \pi_{\mathrm{mix}}(t \mid v) \;=\; \pi_{\mathrm{one}}(t \mid v)
    \;+\; \underbrace{q(d_v)\,\big(\pi_{d_v}(t \mid v) - \pi_{\mathrm{one}}(t \mid v)\big)}_{\text{what reconsideration adds}},
    \label{eq:refprice}
\end{equation}
where the gate's own $q(d_v)$ sets how far the reconsidered distribution is trusted over
the one-pass one, mimicking the weighting the training objective of
Eq.~\eqref{eq:fullloss} optimizes. Reconsideration thus corrects the probability of an edge, upward where a further
pass favors the token and downward where it does not; since $s(\cdot)$ sums edge
log-probabilities along the path, the change $\log\pi_{\mathrm{mix}}-\log\pi_{\mathrm{one}}$ on an edge is added to the score of every node beneath it. A node the
bid never selects has no $\pi_{d_v}$, so Eq.~\eqref{eq:refprice} loses its second term and
returns $\pi_{\mathrm{one}}$. The global cut then ranks reconsidered and one-pass nodes alike, and the target verifies
the final tree. Appendix~\ref{app:serving} gives
the full discussion on serving algorithm of Parallel Reconsideration.

\section{Experiments}
\label{sec:experiments}

We describe the setup (Section~\ref{sec:exp_setup}), then report acceptance length and
speedup against EAGLE-3.1 at a matched budget (Section~\ref{sec:exp_main}), the gain
under batched serving (Section~\ref{sec:exp_batching}), and an ablation of each design
choice and of a larger drafter (Section~\ref{sec:exp_design}). Two further studies appear
in Appendix~\ref{app:experiments}: a comparison against spending the same compute on a
wider draft (Appendix~\ref{app:exp_ablation}), and a measurement of which of the
drafter's failures reconsideration recovers (Appendix~\ref{app:exp_recovery}).

\subsection{Experimental setup}
\label{sec:exp_setup}

We compare \textbf{Parallel Reconsideration} against \textbf{EAGLE-3.1}, the
state-of-the-art feature-level speculative decoder, on three target models,
\textbf{Llama-3.1-8B-Instruct}, \textbf{Qwen3-8B}, and \textbf{Gemma-3-12B}, and across
seven benchmarks spanning math (GSM8K, MATH-500, AIME25), code (HumanEval, MBPP,
LiveCodeBench), and chat (MT-Bench). Both drafters use a single decoder layer and share
an identical target model, both are trained in SpecForge~\citep{li2026specforge} under
EAGLE-3's released recipe, and both are served through
SGLang~\citep{zheng2024sglang} under the same budget configuration, so every number we
report is measured end to end on a production serving engine. Because SGLang already
removes the system bottlenecks that an under-optimized autoregressive baseline
suffers from, the speedups we measure are lower than those reported in the original
EAGLE-3 paper. As an external check on the baseline, our EAGLE-3.1
acceptance lengths agree with those reported independently for the same models and
benchmarks by~\citet{chen2026dflash}. Our primary metrics are the
\emph{acceptance length} $\tau$ and wall-clock speedup over autoregressive
decoding on one H200. Appendix~\ref{app:exp_setup} gives the models, benchmarks,
training and serving details, and the budget configuration in full.

\subsection{Acceptance length and speedup}
\label{sec:exp_main}

\begin{table*}[t]
    \centering
    \caption{\textbf{Acceptance length $\tau$ and decoding speedup over autoregressive
            decoding.} Parenthesized values give the total draft budget $B$ (number of draft
        nodes verified per round), held equal for both methods along with the offer width,
        grow width, and number of draft steps.
        $\Delta$ rows give the relative gain of PR over EAGLE-3.1 at the same budget.
    }
    \label{tab:main_results}
    \resizebox{\linewidth}{!}{
        \begin{tabular}{llcccccccccccccccc}
            \toprule
                           &                  & \multicolumn{6}{c}{\textbf{Math}} & \multicolumn{6}{c}{\textbf{Code}} & \multicolumn{2}{c}{\textbf{Chat}} &                                                                                                                                                                                                                                                                                                                            \\
            \cmidrule(lr){3-8}\cmidrule(lr){9-14}\cmidrule(lr){15-16}
                           &                  & \multicolumn{2}{c}{GSM8K}         & \multicolumn{2}{c}{MATH-500}      & \multicolumn{2}{c}{AIME25}        & \multicolumn{2}{c}{HumanEval} & \multicolumn{2}{c}{MBPP} & \multicolumn{2}{c}{LCB} & \multicolumn{2}{c}{MT-Bench} & \multicolumn{2}{c}{\textbf{Avg.}}                                                                                                                                                                      \\
            \cmidrule(lr){3-4}\cmidrule(lr){5-6}\cmidrule(lr){7-8}\cmidrule(lr){9-10}\cmidrule(lr){11-12}\cmidrule(lr){13-14}\cmidrule(lr){15-16}\cmidrule(lr){17-18}
            \textbf{Model} & \textbf{Method}  & $\tau$                            & Speedup                           & $\tau$                            & Speedup                       & $\tau$                   & Speedup                 & $\tau$                       & Speedup                           & $\tau$        & Speedup               & $\tau$        & Speedup               & $\tau$        & Speedup               & $\tau$             & Speedup               \\
            \midrule
            \multicolumn{18}{l}{\emph{Temperature $=0$}}                                                                                                                                                                                                                                                                                                                                                                                                                               \\
            \midrule
            \multirow{6}{*}{Llama-3.1-8B}
                           & EAGLE-3.1 (24)   & 4.41                              & 1.90$\times$                      & 4.70                              & 2.14$\times$                  & \textbf{5.41}            & \textbf{2.29$\times$}   & 5.15                         & 2.25$\times$                      & 4.97          & 2.24$\times$          & 4.31          & 1.78$\times$          & 4.20          & 1.69$\times$          & 4.74               & 2.04$\times$          \\
                           & \textbf{PR (24)} & \textbf{4.73}                     & \textbf{2.03$\times$}             & \textbf{4.77}                     & \textbf{2.16$\times$}         & 5.36                     & 2.28$\times$            & \textbf{5.45}                & \textbf{2.38$\times$}             & \textbf{5.28} & \textbf{2.37$\times$} & \textbf{4.78} & \textbf{1.95$\times$} & \textbf{4.65} & \textbf{1.85$\times$} & \textbf{5.00}      & \textbf{2.15$\times$} \\
                           & \quad $\Delta$   & $+7.3\%$   & $+6.8\%$   & $+1.5\%$   & $+0.9\%$   & $-0.9\%$   & $-0.4\%$   & $+5.8\%$   & $+5.8\%$   & $+6.2\%$   & $+5.8\%$   & $+10.9\%$   & $+9.6\%$   & $+10.7\%$   & $+9.5\%$   & $\mathbf{+5.5\%}$  & $\mathbf{+5.4\%}$     \\
                           & EAGLE-3.1 (48)   & 4.84                              & 2.05$\times$                      & 4.99                              & 2.13$\times$                  & 5.28                     & 2.27$\times$            & 5.62                         & 2.49$\times$                      & 5.32          & 2.33$\times$          & 4.70          & 1.95$\times$          & 4.60          & 2.03$\times$          & 5.05               & 2.18$\times$          \\
                           & \textbf{PR (48)} & \textbf{5.40}                     & \textbf{2.26$\times$}             & \textbf{5.31}                     & \textbf{2.23$\times$}         & \textbf{5.53}            & \textbf{2.37$\times$}   & \textbf{5.87}                & \textbf{2.56$\times$}             & \textbf{5.57} & \textbf{2.41$\times$} & \textbf{5.11} & \textbf{2.11$\times$} & \textbf{4.77} & \textbf{2.05$\times$} & \textbf{5.37}      & \textbf{2.28$\times$} \\
                           & \quad $\Delta$   & $+11.6\%$   & $+10.2\%$   & $+6.4\%$   & $+4.7\%$   & $+4.7\%$   & $+4.4\%$   & $+4.4\%$   & $+2.8\%$   & $+4.7\%$   & $+3.4\%$   & $+8.7\%$   & $+8.2\%$   & $+3.7\%$   & $+1.0\%$   & $\mathbf{+6.3\%}$  & $\mathbf{+4.6\%}$     \\
            \cmidrule(lr){1-18}
            \multirow{6}{*}{Qwen3-8B}
                           & EAGLE-3.1 (24)   & 4.39                              & 1.94$\times$                      & 3.96                              & 2.04$\times$                  & 3.81                     & 1.86$\times$            & 4.46                         & 2.11$\times$                      & 4.34          & 2.11$\times$          & 3.66          & 1.78$\times$          & 3.36          & 1.57$\times$          & 4.00               & 1.92$\times$          \\
                           & \textbf{PR (24)} & \textbf{4.83}                     & \textbf{2.11$\times$}             & \textbf{4.26}                     & \textbf{2.16$\times$}         & \textbf{4.20}            & \textbf{2.00$\times$}   & \textbf{4.79}                & \textbf{2.26$\times$}             & \textbf{4.84} & \textbf{2.32$\times$} & \textbf{3.99} & \textbf{1.91$\times$} & \textbf{3.59} & \textbf{1.66$\times$} & \textbf{4.36}      & \textbf{2.06$\times$} \\
                           & \quad $\Delta$   & $+10.0\%$   & $+8.8\%$   & $+7.6\%$   & $+5.9\%$   & $+10.2\%$   & $+7.5\%$   & $+7.4\%$   & $+7.1\%$   & $+11.5\%$   & $+10.0\%$   & $+9.0\%$   & $+7.3\%$   & $+6.8\%$   & $+5.7\%$   & $\mathbf{+9.0\%}$  & $\mathbf{+7.3\%}$     \\
                           & EAGLE-3.1 (48)   & 4.84                              & 2.18$\times$                      & 4.30                              & 2.24$\times$                  & 4.23                     & 1.99$\times$            & 4.85                         & 2.41$\times$                      & 4.75          & 2.22$\times$          & 4.00          & 1.86$\times$          & 3.66          & 1.71$\times$          & 4.38               & 2.09$\times$          \\
                           & \textbf{PR (48)} & \textbf{5.13}                     & \textbf{2.29$\times$}             & \textbf{4.50}                     & \textbf{2.32$\times$}         & \textbf{4.60}            & \textbf{2.10$\times$}   & \textbf{5.03}                & \textbf{2.50$\times$}             & \textbf{5.07} & \textbf{2.33$\times$} & \textbf{4.18} & \textbf{1.93$\times$} & \textbf{3.82} & \textbf{1.77$\times$} & \textbf{4.62}      & \textbf{2.18$\times$} \\
                           & \quad $\Delta$   & $+6.0\%$   & $+5.0\%$   & $+4.7\%$   & $+3.6\%$   & $+8.7\%$   & $+5.5\%$   & $+3.7\%$   & $+3.7\%$   & $+6.7\%$   & $+5.0\%$   & $+4.5\%$   & $+3.8\%$   & $+4.4\%$   & $+3.5\%$   & $\mathbf{+5.5\%}$  & $\mathbf{+4.3\%}$     \\
            \cmidrule(lr){1-18}
            \multirow{6}{*}{Gemma-3-12B}
                           & EAGLE-3.1 (24)   & 4.92                              & 3.08$\times$                      & 4.21                              & 2.54$\times$                  & 4.29                     & 2.68$\times$            & 6.37                         & 3.40$\times$                      & 5.29          & 3.08$\times$          & 5.28          & 2.78$\times$          & 3.25          & 1.79$\times$          & 4.80               & 2.76$\times$          \\
                           & \textbf{PR (24)} & \textbf{5.28}                     & \textbf{3.16$\times$}             & \textbf{4.65}                     & \textbf{2.78$\times$}         & \textbf{4.60}            & \textbf{2.86$\times$}   & \textbf{7.07}                & \textbf{3.78$\times$}             & \textbf{5.65} & \textbf{3.30$\times$} & \textbf{5.90} & \textbf{3.05$\times$} & \textbf{3.75} & \textbf{1.98$\times$} & \textbf{5.27}      & \textbf{2.99$\times$} \\
                           & \quad $\Delta$   & $+7.3\%$   & $+2.6\%$   & $+10.5\%$   & $+9.4\%$   & $+7.2\%$   & $+6.7\%$   & $+11.0\%$   & $+11.2\%$   & $+6.8\%$   & $+7.1\%$   & $+11.7\%$   & $+9.7\%$   & $+15.4\%$   & $+10.6\%$   & $\mathbf{+9.8\%}$  & $\mathbf{+8.3\%}$     \\
                           & EAGLE-3.1 (48)   & 5.43                              & 3.23$\times$                      & 4.53                              & 2.60$\times$                  & 4.67                     & 2.93$\times$            & 6.73                         & 3.68$\times$                      & 5.68          & 3.20$\times$          & 5.81          & 3.27$\times$          & 3.58          & 1.94$\times$          & 5.20               & 2.98$\times$          \\
                           & \textbf{PR (48)} & \textbf{5.96}                     & \textbf{3.53$\times$}             & \textbf{4.89}                     & \textbf{2.80$\times$}         & \textbf{4.93}            & \textbf{3.07$\times$}   & \textbf{7.10}                & \textbf{3.84$\times$}             & \textbf{6.08} & \textbf{3.33$\times$} & \textbf{6.46} & \textbf{3.60$\times$} & \textbf{3.94} & \textbf{2.09$\times$} & \textbf{5.62}      & \textbf{3.18$\times$} \\
                           & \quad $\Delta$   & $+9.8\%$   & $+9.3\%$   & $+7.9\%$   & $+7.7\%$   & $+5.6\%$   & $+4.8\%$   & $+5.5\%$   & $+4.3\%$   & $+7.0\%$   & $+4.1\%$   & $+11.2\%$   & $+10.1\%$   & $+10.1\%$   & $+7.7\%$   & $\mathbf{+8.1\%}$  & $\mathbf{+6.7\%}$     \\
            \midrule
            \multicolumn{18}{l}{\emph{Temperature $=1$}}                                                                                                                                                                                                                                                                                                                                                                                                                               \\
            \midrule
            \multirow{6}{*}{Llama-3.1-8B}
                           & EAGLE-3.1 (24)   & 3.47                              & 1.45$\times$                      & 4.13                              & 2.02$\times$                  & 4.49                     & 2.06$\times$            & 4.61                         & 2.16$\times$                      & 4.40          & 1.99$\times$          & 2.82          & 1.21$\times$          & 2.07          & 1.17$\times$          & 3.71               & 1.72$\times$          \\
                           & \textbf{PR (24)} & \textbf{4.01}                     & \textbf{1.61$\times$}             & \textbf{4.45}                     & \textbf{2.12$\times$}         & \textbf{4.51}            & \textbf{2.08$\times$}   & \textbf{4.80}                & \textbf{2.18$\times$}             & \textbf{4.55} & \textbf{2.03$\times$} & \textbf{3.15} & \textbf{1.32$\times$} & \textbf{2.29} & \textbf{1.24$\times$} & \textbf{3.97}      & \textbf{1.80$\times$} \\
                           & \quad $\Delta$   & $+15.6\%$   & $+11.0\%$   & $+7.7\%$    & $+5.0\%$    & $+0.4\%$   & $+1.0\%$   & $+4.1\%$   & $+0.9\%$   & $+3.4\%$   & $+2.0\%$   & $+11.7\%$   & $+9.1\%$   & $+10.6\%$   & $+6.0\%$   & $\mathbf{+7.0\%}$  & $\mathbf{+4.7\%}$     \\
                           & EAGLE-3.1 (48)   & 3.99                              & 1.67$\times$                      & 4.44                              & 2.06$\times$                  & 4.51                     & 2.13$\times$            & 5.13                         & 2.31$\times$                      & 4.79          & 2.16$\times$          & 3.02          & 1.29$\times$          & 2.72          & 1.45$\times$          & 4.09               & 1.87$\times$          \\
                           & \textbf{PR (48)} & \textbf{4.56}                     & \textbf{1.84$\times$}             & \textbf{4.79}                     & \textbf{2.19$\times$}         & \textbf{4.58}            & \textbf{2.16$\times$}   & \textbf{5.42}                & \textbf{2.36$\times$}             & \textbf{5.08} & \textbf{2.23$\times$} & \textbf{3.49} & \textbf{1.47$\times$} & \textbf{2.83} & \textbf{1.49$\times$} & \textbf{4.39}      & \textbf{1.96$\times$} \\
                           & \quad $\Delta$   & $+14.3\%$   & $+10.2\%$   & $+7.9\%$    & $+6.3\%$   & $+1.6\%$   & $+1.4\%$   & $+5.7\%$   & $+2.2\%$   & $+6.1\%$   & $+3.2\%$   & $+15.6\%$   & $+14.0\%$   & $+4.0\%$   & $+2.8\%$   & $\mathbf{+7.3\%}$  & $\mathbf{+4.8\%}$     \\
            \cmidrule(lr){1-18}
            \multirow{6}{*}{Qwen3-8B}
                           & EAGLE-3.1 (24)   & 4.47                              & 2.17$\times$                      & 3.62                              & 1.74$\times$                  & 3.25                     & 1.52$\times$            & 4.32                         & 1.94$\times$                      & 4.08          & 1.78$\times$          & 3.38          & 1.52$\times$          & 3.12          & 1.47$\times$          & 3.75               & 1.73$\times$          \\
                           & \textbf{PR (24)} & \textbf{4.76}                     & \textbf{2.27$\times$}             & \textbf{3.90}                     & \textbf{1.84$\times$}         & \textbf{3.43}            & \textbf{1.57$\times$}   & \textbf{4.50}                & \textbf{1.96$\times$}             & \textbf{4.30} & \textbf{1.84$\times$} & \textbf{3.53} & \textbf{1.57$\times$} & \textbf{3.25} & \textbf{1.50$\times$} & \textbf{3.95}      & \textbf{1.79$\times$} \\
                           & \quad $\Delta$   & $+6.5\%$   & $+4.6\%$   & $+7.7\%$   & $+5.7\%$   & $+5.5\%$   & $+3.3\%$   & $+4.2\%$   & $+1.0\%$   & $+5.4\%$   & $+3.4\%$   & $+4.4\%$   & $+3.3\%$   & $+4.2\%$   & $+2.0\%$   & $\mathbf{+5.3\%}$  & $\mathbf{+3.5\%}$     \\
                           & EAGLE-3.1 (48)   & 4.84                              & 2.20$\times$                      & 3.94                              & 1.94$\times$                  & 3.59                     & 1.70$\times$            & 4.73                         & 2.26$\times$                      & 4.48          & 2.15$\times$          & 3.67          & 1.70$\times$          & 3.37          & 1.57$\times$          & 4.09               & 1.93$\times$          \\
                           & \textbf{PR (48)} & \textbf{5.21}                     & \textbf{2.36$\times$}             & \textbf{4.19}                     & \textbf{2.00$\times$}         & \textbf{3.70}            & \textbf{1.72$\times$}   & \textbf{4.95}                & \textbf{2.40$\times$}             & \textbf{4.69} & \textbf{2.24$\times$} & \textbf{3.92} & \textbf{1.78$\times$} & \textbf{3.63} & \textbf{1.65$\times$} & \textbf{4.33}      & \textbf{2.02$\times$} \\
                           & \quad $\Delta$   & $+7.6\%$   & $+7.3\%$   & $+6.3\%$   & $+3.1\%$   & $+3.1\%$   & $+1.2\%$   & $+4.7\%$   & $+6.2\%$   & $+4.7\%$   & $+4.2\%$   & $+6.8\%$   & $+4.7\%$   & $+7.7\%$   & $+5.1\%$   & $\mathbf{+5.9\%}$  & $\mathbf{+4.7\%}$     \\
            \cmidrule(lr){1-18}
            \multirow{6}{*}{Gemma-3-12B}
                           & EAGLE-3.1 (24)   & 5.00                              & 2.79$\times$                      & 4.12                              & 2.41$\times$                  & 3.79                     & 2.00$\times$            & 6.41                         & 3.33$\times$                      & 5.44          & 3.17$\times$          & 5.17          & 3.17$\times$          & 3.12          & 1.48$\times$          & 4.72               & 2.62$\times$          \\
                           & \textbf{PR (24)} & \textbf{5.38}                     & \textbf{2.90$\times$}             & \textbf{4.38}                     & \textbf{2.54$\times$}         & \textbf{4.29}            & \textbf{2.24$\times$}   & \textbf{7.51}                & \textbf{3.86$\times$}             & \textbf{5.83} & \textbf{3.32$\times$} & \textbf{5.59} & \textbf{3.41$\times$} & \textbf{3.28} & \textbf{1.54$\times$} & \textbf{5.18}      & \textbf{2.83$\times$} \\
                           & \quad $\Delta$   & $+7.6\%$   & $+3.9\%$   & $+6.3\%$   & $+5.4\%$   & $+13.2\%$   & $+12.0\%$   & $+17.2\%$   & $+15.9\%$   & $+7.2\%$   & $+4.7\%$   & $+8.1\%$   & $+7.6\%$   & $+5.1\%$   & $+4.1\%$   & $\mathbf{+9.7\%}$  & $\mathbf{+8.0\%}$     \\
                           & EAGLE-3.1 (48)   & 5.44                              & 2.81$\times$                      & 4.47                              & 2.36$\times$                  & 4.35                     & 2.14$\times$            & 6.70                         & 3.89$\times$                      & 5.83          & 2.98$\times$          & 5.54          & 3.13$\times$          & 3.40          & 1.64$\times$          & 5.10               & 2.71$\times$          \\
                           & \textbf{PR (48)} & \textbf{6.22}                     & \textbf{3.19$\times$}             & \textbf{4.82}                     & \textbf{2.53$\times$}         & \textbf{4.86}            & \textbf{2.33$\times$}   & \textbf{7.25}                & \textbf{4.20$\times$}             & \textbf{6.66} & \textbf{3.39$\times$} & \textbf{5.99} & \textbf{3.37$\times$} & \textbf{3.55} & \textbf{1.70$\times$} & \textbf{5.62}      & \textbf{2.96$\times$} \\
                           & \quad $\Delta$   & $+14.3\%$   & $+13.5\%$   & $+7.8\%$   & $+7.2\%$   & $+11.7\%$   & $+8.9\%$   & $+8.2\%$   & $+8.0\%$   & $+14.2\%$   & $+13.8\%$   & $+8.1\%$   & $+7.7\%$   & $+4.4\%$   & $+3.7\%$   & $\mathbf{+10.2\%}$  & $\mathbf{+9.2\%}$     \\
            \bottomrule
        \end{tabular}
    }
\end{table*}

\begin{wraptable}{r}{0.5\linewidth}
    \vspace{-1.2em}
    \centering
    \scriptsize
    \setlength{\tabcolsep}{2.5pt}
    \caption{\textbf{Speedup as requests are batched.}
        Wall-clock speedup of EAGLE-3.1 and Parallel Reconsideration (PR) over autoregressive decoding at a
        matched batch size $n$, measured on the same serving engine ($B=48$, temperature $0$).}
    \label{tab:batching}
    \resizebox{\linewidth}{!}{
        \begin{tabular}{lllccccc}
            \toprule
                  &          &             & \multicolumn{5}{c}{\textbf{Batch size $n$}}                                                             \\
            \cmidrule(lr){4-8}
                  &          &             & \textbf{1}                                  & \textbf{2}   & \textbf{4}   & \textbf{8}   & \textbf{16}  \\
            \midrule
            Llama & GSM8K    & EAGLE-3.1   & 2.05$\times$                                & 1.68$\times$ & 1.62$\times$ & 1.26$\times$ & 1.12$\times$ \\
                  &          & \textbf{PR} & 2.26$\times$                                & 1.91$\times$ & 1.91$\times$ & 1.46$\times$ & 1.22$\times$ \\
                  & MT-Bench & EAGLE-3.1   & 2.03$\times$                                & 1.71$\times$ & 1.71$\times$ & 1.50$\times$ & 1.41$\times$ \\
                  &          & \textbf{PR} & 2.05$\times$                                & 1.79$\times$ & 1.77$\times$ & 1.56$\times$ & 1.51$\times$ \\
                  & LCB      & EAGLE-3.1   & 1.95$\times$                                & 1.61$\times$ & 1.60$\times$ & 1.40$\times$ & 1.31$\times$ \\
                  &          & \textbf{PR} & 2.11$\times$                                & 1.84$\times$ & 1.78$\times$ & 1.33$\times$ & 1.29$\times$ \\
            \cmidrule(lr){1-8}
            Qwen  & GSM8K    & EAGLE-3.1   & 2.18$\times$                                & 2.16$\times$ & 2.01$\times$ & 1.68$\times$ & 1.39$\times$ \\
                  &          & \textbf{PR} & 2.29$\times$                                & 2.26$\times$ & 2.25$\times$ & 1.94$\times$ & 1.64$\times$ \\
                  & MT-Bench & EAGLE-3.1   & 1.71$\times$                                & 1.72$\times$ & 1.71$\times$ & 1.35$\times$ & 1.11$\times$ \\
                  &          & \textbf{PR} & 1.77$\times$                                & 1.77$\times$ & 1.69$\times$ & 1.41$\times$ & 1.24$\times$ \\
                  & LCB      & EAGLE-3.1   & 1.86$\times$                                & 1.81$\times$ & 1.80$\times$ & 1.60$\times$ & 1.31$\times$ \\
                  &          & \textbf{PR} & 1.93$\times$                                & 1.87$\times$ & 1.95$\times$ & 1.66$\times$ & 1.43$\times$ \\
            \cmidrule(lr){1-8}
            Gemma & GSM8K    & EAGLE-3.1   & 3.23$\times$                                & 2.79$\times$ & 2.54$\times$ & 2.50$\times$ & 1.71$\times$ \\
                  &          & \textbf{PR} & 3.53$\times$                                & 2.99$\times$ & 2.72$\times$ & 2.70$\times$ & 1.94$\times$ \\
                  & MT-Bench & EAGLE-3.1   & 1.94$\times$                                & 1.92$\times$ & 1.83$\times$ & 1.71$\times$ & 1.50$\times$ \\
                  &          & \textbf{PR} & 2.09$\times$                                & 2.02$\times$ & 1.91$\times$ & 1.91$\times$ & 1.40$\times$ \\
                  & LCB      & EAGLE-3.1   & 3.27$\times$                                & 2.79$\times$ & 2.72$\times$ & 2.55$\times$ & 2.04$\times$ \\
                  &          & \textbf{PR} & 3.60$\times$                                & 3.02$\times$ & 2.97$\times$ & 2.85$\times$ & 2.14$\times$ \\
            \bottomrule
        \end{tabular}
    }
    \vspace{-1em}
\end{wraptable}

Table~\ref{tab:main_results} reports acceptance length and speedup across all seven
benchmarks. Parallel Reconsideration improves average acceptance length over EAGLE-3.1 at
every budget, on all three models, and at both temperatures, by \textbf{5.3--10.2\%}, and
the gain carries through to wall-clock speedup (\textbf{3.5--9.2\%}) at a nearly
identical per-round drafting cost (Appendix~\ref{app:coforward}). The gain is consistent across benchmarks from different domains: LiveCodeBench reaches
\textbf{+10.1\%} speedup on Gemma-3, and \textbf{+10.2\%} speedup on GSM8K for Llama3.1.

\subsection{Serving at larger batch sizes}
\label{sec:exp_batching}

Speculative decoding is hardest to justify at large batch sizes, where the target's
forward is already compute-bound and every drafted node competes for the same
verification capacity. We therefore benchmark Parallel Reconsideration at batch sizes
$n \in \{1, 2, 4, 8, 16\}$ against autoregressive decoding at the same $n$
(Table~\ref{tab:batching}).

Gemma-3-12B holds \textbf{2.85$\times$} on
LiveCodeBench at $n=8$ and \textbf{2.14$\times$} at $n=16$, and Qwen3-8B is flat through
$n=4$ (\textbf{2.25$\times$} on GSM8K, \textbf{1.95$\times$} on LiveCodeBench). Reconsideration buys acceptance out of
the draft budget rather than the verification budget
(Section~\ref{sec:serving}), so the target still verifies the same $B$ nodes per round at
every $n$, and the co-forwarded rows are absorbed by the drafter's own batch rather than
the target's.

\subsection{Each component, and the alternative of a bigger drafter}
\label{sec:exp_design}

\begin{table}[t]
    \centering
    \caption{\textbf{Removing each design choice, and spending the compute on depth instead.}
        Llama-3.1-8B at $B=48$, temperature $0$, reporting $\tau$ / speedup. The first two
        rows remove one component of Parallel Reconsideration and retrain; the third is
        EAGLE-3.1 with a second decoder layer and no reconsideration, spending its
        additional capacity on every position rather than on the uncertain ones.
        $\Delta$ is relative to Parallel Reconsideration.}
    \label{tab:ablation_design}
    \resizebox{\linewidth}{!}{
        \begin{tabular}{lccccccccc}
            \toprule
                                               & \multicolumn{2}{c}{GSM8K} & \multicolumn{2}{c}{LiveCodeBench} & \multicolumn{2}{c}{MBPP} & \multicolumn{2}{c}{\textbf{Avg.}}                                                                                 \\
            \cmidrule(lr){2-3}\cmidrule(lr){4-5}\cmidrule(lr){6-7}\cmidrule(lr){8-9}
            \textbf{Drafter}                   & $\tau$                    & Speedup                           & $\tau$                   & Speedup                           & $\tau$        & Speedup               & $\tau$        & Speedup               \\
            \midrule
            EAGLE-3.1                          & 4.84                      & 2.05$\times$                      & 4.70                     & 1.95$\times$                      & 5.32          & 2.33$\times$          & 4.95          & 2.11$\times$          \\
            \textbf{Parallel Reconsideration}  & \textbf{5.40}             & \textbf{2.26$\times$}             & \textbf{5.11}            & \textbf{2.11$\times$}             & \textbf{5.57} & \textbf{2.41$\times$} & \textbf{5.36} & \textbf{2.26$\times$} \\
            \midrule
            \quad no balance term ($\beta=0$)  & 5.17                      & 2.14$\times$                      & 5.03                     & 2.04$\times$                      & 5.58          & 2.41$\times$          & 5.26          & 2.20$\times$          \\
            \quad\quad $\Delta$                & $-4.3\%$                  & $-5.3\%$                          & $-1.6\%$                 & $-3.3\%$                          & $+0.2\%$      & $+0.0\%$              & $-1.9\%$      & $-2.8\%$              \\
            \quad no self-query ($Q \to e(t_p)$) & 5.03                      & 2.10$\times$                      & 4.92                     & 2.02$\times$                      & 5.47          & 2.35$\times$          & 5.14          & 2.16$\times$          \\
            \quad\quad $\Delta$                & $-6.9\%$                  & $-7.1\%$                          & $-3.7\%$                 & $-4.3\%$                          & $-1.8\%$      & $-2.5\%$              & $-4.1\%$      & $-4.6\%$              \\
            \midrule
            EAGLE-3.1, two decoder layers      & 4.87                      & 1.90$\times$                      & 4.91                     & 1.85$\times$                      & 5.37          & 2.14$\times$          & 5.05          & 1.96$\times$          \\
            \quad\quad $\Delta$                & $-9.8\%$                  & $\mathbf{-15.9\%}$                & $-3.9\%$                 & $\mathbf{-12.3\%}$                & $-3.6\%$      & $\mathbf{-11.2\%}$    & $-5.8\%$      & $\mathbf{-13.1\%}$    \\
            \bottomrule
        \end{tabular}
    }
\end{table}

Table~\ref{tab:ablation_design} removes one design choice at a time and retrains. It
also answers a question the cost model raises: if reconsideration is worth a second pass
of $M$, why not simply give the drafter a second layer and spend that capacity on every
position?

\textbf{The self-query.} Replacing $Q(f_p^{0})$ with $e(t_p)$, the embedding of the token
that produced $f_p^{0}$, costs $4.1\%$ of average acceptance and $4.6\%$ of speedup.
Steering reconsideration with a committed
token rather than the feature's own query recovers less, which is what
Section~\ref{sec:method_resolve} predicts: a token embedding directs $M$ to advance
along that branch, while reconsideration needs to stay in place.

\textbf{The balance term.} Setting $\beta=0$ costs $1.9\%$ of average acceptance. The term keeps a depth from collapse before the head's term can train it, so removing it degrades the drafter without disabling
it.

\textbf{A bigger drafter instead.} Giving EAGLE-3.1 a second decoder layer raises its
acceptance, by $2.0\%$ on average over the one-layer EAGLE baseline, and still \emph{loses}
$7.0\%$ of its speedup, because the extra layer raises $t_{\mathrm{step}}$ on every
draft step. Against Parallel Reconsideration
the gap is \textbf{13.1\%} of speedup at $5.8\%$ lower acceptance.


\section{Related Work}
\label{sec:related_work}

Speculative decoding accelerates autoregressive generation by drafting several tokens
cheaply and verifying them in one target forward pass, accepting the longest prefix the
target agrees with so the output distribution is preserved exactly
\citep{leviathan2023fast, xia2023speculative}; since the acceptance rule guarantees that
distribution, subsequent work buys speed either by raising the \emph{acceptance length}
or by lowering the \emph{drafting cost}. Self-drafting methods derive the draft from the
target's own feature, as in the EAGLE family, which autoregresses on target features
with a single lightweight decoder layer \citep{li2024eagle, li2024eagle2, li2025eagle3,
    eldenk2026attention}, or in the prediction heads attached to the target's last hidden
state \citep{cai2024medusa, ankner2024hydra}. Training-free drafters remove the auxiliary
model altogether and verify $n$-gram continuations generated in parallel
\citep{fu2024lookahead, le2026fafo}, while parallel drafters attack the cost axis
directly by emitting an entire block per forward pass \citep{an2025pard, chen2026dflash,
    cheng2026dspark}. A separate line changes what the drafter optimizes, recasting draft
training as variational inference over draft paths so that acceptance, rather than token
likelihood, is the objective \citep{zou2026vsd}. Parallel Reconsideration is orthogonal
to most of these: it adds a second guess at the uncertain positions of an existing
one-pass draft, and applies to any drafter that exposes a reusable draft feature.
Finally, spending extra passes to sharpen a prediction is itself familiar, from looped
and universal transformers to adaptive-compute methods that learn how many steps each
token needs \citep{giannou2023looped, dehghani2019universal, banino2021pondernet,
    schuster2022calm, zhu2025ouro}; we loop not to make an autoregressive model reason
better, but to revise the failure of autoregressive \emph{drafting}, and we run
the extra pass in parallel so it costs no additional step.
Appendix~\ref{app:related_work} provides a more detailed discussion of related work.

\section{Conclusion}
\label{sec:conclusion}

We presented \textbf{Parallel Reconsideration}, which grants the drafter a \emph{second
    chance to guess} on the tokens it is unsure about: where it is already decisive the
guess is committed as usual, and where it hesitates the drafter is given a few extra
attempts to revise it. The attempts are batched into the same forward pass as the
ordinary draft, so they add no sequential draft step and cost only negligible parallel
overhead. This brings \emph{draft-time scaling} to feature-level drafters, letting the
drafter vary how much it computes per position rather than how well it computes in one
step. Across Llama-3.1-8B-Instruct, Qwen3-8B, and Gemma-3-12B, it raises average
acceptance length over EAGLE-3.1 by \textbf{5.3--10.2\%} and wall-clock speedup by
\textbf{3.5--9.2\%} on math, code, and chat benchmarks at a matched draft budget.

\bibliography{main}

@misc{anthropic_claudecode,
  title        = {Claude Code},
  author       = {{Anthropic}},
  year         = {2025},
  howpublished = {\url{https://claude.com/product/claude-code}}
}

@misc{openai_codex,
  title        = {Codex},
  author       = {{OpenAI}},
  year         = {2025},
  howpublished = {\url{https://openai.com/codex/}}
}

@misc{google_antigravity,
  title        = {Google Antigravity},
  author       = {{Google}},
  year         = {2025},
  howpublished = {\url{https://antigravity.google/}}
}

@misc{moonshot_kimik3,
  title        = {{Kimi K3}},
  author       = {{Moonshot AI}},
  year         = {2026},
  howpublished = {\url{https://www.kimi.ai/blog/kimi-k3}}
}

@inproceedings{pope2023efficiently,
  title     = {Efficiently Scaling Transformer Inference},
  author    = {Pope, Reiner and Douglas, Sholto and Chowdhery, Aakanksha and Devlin, Jacob and Bradbury, James and Levskaya, Anselm and Heek, Jonathan and Xiao, Kefan and Agrawal, Shivani and Dean, Jeff},
  booktitle = {Proceedings of Machine Learning and Systems (MLSys)},
  volume    = {5},
  pages     = {606--624},
  year      = {2023}
}

@inproceedings{leviathan2023fast,
  title        = {Fast Inference from Transformers via Speculative Decoding},
  author       = {Leviathan, Yaniv and Kalman, Matan and Matias, Yossi},
  booktitle    = {International Conference on Machine Learning (ICML)},
  pages        = {19274--19286},
  year         = {2023},
  organization = {PMLR}
}

@inproceedings{xia2023speculative,
  title     = {Speculative Decoding: Exploiting Speculative Execution for Accelerating Seq2seq Generation},
  author    = {Xia, Heming and Ge, Tao and Wang, Peiyi and Chen, Si-Qing and Wei, Furu and Sui, Zhifang},
  booktitle = {Findings of the Association for Computational Linguistics: EMNLP 2023},
  pages     = {3909--3925},
  year      = {2023}
}

@inproceedings{li2024eagle,
  title        = {{EAGLE}: Speculative Sampling Requires Rethinking Feature Uncertainty},
  author       = {Li, Yuhui and Wei, Fangyun and Zhang, Chao and Zhang, Hongyang},
  booktitle    = {International Conference on Machine Learning (ICML)},
  pages        = {28935--28948},
  year         = {2024},
  organization = {PMLR}
}

@inproceedings{li2024eagle2,
  title     = {{EAGLE-2}: Faster Inference of Language Models with Dynamic Draft Trees},
  author    = {Li, Yuhui and Wei, Fangyun and Zhang, Chao and Zhang, Hongyang},
  booktitle = {Proceedings of the 2024 Conference on Empirical Methods in Natural Language Processing (EMNLP)},
  pages     = {7421--7432},
  year      = {2024}
}

@inproceedings{li2025eagle3,
  title     = {{EAGLE-3}: Scaling up Inference Acceleration of Large Language Models via Training-Time Test},
  author    = {Li, Yuhui and Wei, Fangyun and Zhang, Chao and Zhang, Hongyang},
  booktitle = {Advances in Neural Information Processing Systems (NeurIPS)},
  year      = {2025}
}

@article{cobbe2021gsm8k,
  title   = {Training Verifiers to Solve Math Word Problems},
  author  = {Cobbe, Karl and Kosaraju, Vineet and Bavarian, Mohammad and Chen, Mark and Jun, Heewoo and Kaiser, Lukasz and Plappert, Matthias and Tworek, Jerry and Hilton, Jacob and Nakano, Reiichiro and Hesse, Christopher and Schulman, John},
  journal = {arXiv preprint arXiv:2110.14168},
  year    = {2021}
}

@article{lightman2024math500,
  title   = {Let's Verify Step by Step},
  author  = {Lightman, Hunter and Kosaraju, Vineet and Burda, Yura and Edwards, Harrison and Baker, Bowen and Lee, Teddy and Leike, Jan and Schulman, John and Sutskever, Ilya and Cobbe, Karl},
  journal = {arXiv preprint arXiv:2305.20050},
  year    = {2023}
}

@misc{maa2025aime,
  title        = {American Invitational Mathematics Examination ({AIME})},
  author       = {{Mathematical Association of America}},
  year         = {2025},
  howpublished = {\url{https://maa.org/math-competitions/american-invitational-mathematics-examination-aime}}
}

@article{chen2021humaneval,
  title   = {Evaluating Large Language Models Trained on Code},
  author  = {Chen, Mark and Tworek, Jerry and Jun, Heewoo and Yuan, Qiming and de Oliveira Pinto, Henrique Ponde and Kaplan, Jared and Edwards, Harri and Burda, Yuri and Joseph, Nicholas and Brockman, Greg and others},
  journal = {arXiv preprint arXiv:2107.03374},
  year    = {2021}
}

@article{austin2021mbpp,
  title   = {Program Synthesis with Large Language Models},
  author  = {Austin, Jacob and Odena, Augustus and Nye, Maxwell and Bosma, Maarten and Michalewski, Henryk and Dohan, David and Jiang, Ellen and Cai, Carrie and Terry, Michael and Le, Quoc and Sutton, Charles},
  journal = {arXiv preprint arXiv:2108.07732},
  year    = {2021}
}

@inproceedings{jain2025livecodebench,
  title     = {{LiveCodeBench}: Holistic and Contamination Free Evaluation of Large Language Models for Code},
  author    = {Jain, Naman and Han, King and Gu, Alex and Li, Wen-Ding and Yan, Fanjia and Zhang, Tianjun and Wang, Sida and Solar-Lezama, Armando and Sen, Koushik and Stoica, Ion},
  booktitle = {International Conference on Learning Representations (ICLR)},
  year      = {2025}
}

@inproceedings{zheng2023mtbench,
  title     = {Judging {LLM}-as-a-Judge with {MT-Bench} and Chatbot Arena},
  author    = {Zheng, Lianmin and Chiang, Wei-Lin and Sheng, Ying and Zhuang, Siyuan and Wu, Zhanghao and Zhuang, Yonghao and Lin, Zi and Li, Zhuohan and Li, Dacheng and Xing, Eric P. and Zhang, Hao and Gonzalez, Joseph E. and Stoica, Ion},
  booktitle = {Advances in Neural Information Processing Systems (NeurIPS)},
  year      = {2023}
}

@inproceedings{ding2023ultrachat,
  title     = {Enhancing Chat Language Models by Scaling High-quality Instructional Conversations},
  author    = {Ding, Ning and Chen, Yulin and Xu, Bokai and Qin, Yujia and Zheng, Zhi and Hu, Shengding and Liu, Zhiyuan and Sun, Maosong and Zhou, Bowen},
  booktitle = {Proceedings of the 2023 Conference on Empirical Methods in Natural Language Processing (EMNLP)},
  year      = {2023}
}

@article{li2026specforge,
  title   = {{SpecForge}: A Flexible and Efficient Open-Source Training Framework for Speculative Decoding},
  author  = {Li, Shenggui and Wang, Chao and Zhu, Yikai and Wang, Yubo and Yin, Fan and Shi, Shuai and Chen, Yefei and Dong, Xiaomin and Chen, Qiaoling and Pan, Jin and others},
  journal = {arXiv preprint arXiv:2603.18567},
  year    = {2026}
}

@inproceedings{zheng2024sglang,
  title     = {{SGLang}: Efficient Execution of Structured Language Model Programs},
  author    = {Zheng, Lianmin and Yin, Liangsheng and Xie, Zhiqiang and Sun, Chuyue and Huang, Jeff and Yu, Cody Hao and Cao, Shiyi and Kozyrakis, Christos and Stoica, Ion and Gonzalez, Joseph E. and Barrett, Clark and Sheng, Ying},
  booktitle = {Advances in Neural Information Processing Systems (NeurIPS)},
  year      = {2024}
}

@article{eldenk2026attention,
  title   = {Attention Drift: What Autoregressive Speculative Decoding Models Learn},
  author  = {Eldenk, Do{\u{g}}a{\c{c}} and Mohapatra, Payal and Comlek, Yigitcan and Oktay, Kaan and Zhang, Hongyang and Xia, Stephen},
  journal = {arXiv preprint arXiv:2605.09992},
  year    = {2026}
}

@inproceedings{cai2024medusa,
  title        = {{Medusa}: Simple {LLM} Inference Acceleration Framework with Multiple Decoding Heads},
  author       = {Cai, Tianle and Li, Yuhong and Geng, Zhengyang and Peng, Hongwu and Lee, Jason D. and Chen, Deming and Dao, Tri},
  booktitle    = {International Conference on Machine Learning (ICML)},
  year         = {2024},
  organization = {PMLR}
}

@article{ankner2024hydra,
  title   = {{Hydra}: Sequentially-Dependent Draft Heads for {Medusa} Decoding},
  author  = {Ankner, Zachary and Parthasarathy, Rishab and Nrusimha, Aniruddha and Rinard, Christopher and Ragan-Kelley, Jonathan and Brandon, William},
  journal = {arXiv preprint arXiv:2402.05109},
  year    = {2024}
}

@inproceedings{fu2024lookahead,
  title        = {Break the Sequential Dependency of {LLM} Inference Using Lookahead Decoding},
  author       = {Fu, Yichao and Bailis, Peter and Stoica, Ion and Zhang, Hao},
  booktitle    = {International Conference on Machine Learning (ICML)},
  year         = {2024},
  organization = {PMLR}
}

@inproceedings{le2026fafo,
  title        = {{FAFO}: Lossy {KV} Cache Compression for Lossless Inference Acceleration via Draftless Fumble Decoding},
  author       = {Le, Hoang Anh Duy and Zhong, Shaochen and Lu, Yifan and Dou, Yingtong and Yuan, Jiayi and Chuang, Yu-Neng and Fan, Xiran and Wang, Guanchu and Chen, Yuzhong and Hu, Xia},
  booktitle    = {International Conference on Machine Learning (ICML)},
  year         = {2026},
  organization = {PMLR}
}

@article{an2025pard,
  title   = {{PARD}: Accelerating {LLM} Inference with Low-Cost {PAR}allel Draft Model Adaptation},
  author  = {An, Zihao and Bai, Huajun and Liu, Ziqiong and Li, Dong and Barsoum, Emad},
  journal = {arXiv preprint arXiv:2504.18583},
  year    = {2025}
}

@inproceedings{chen2026dflash,
  title     = {{DFlash}: Block Diffusion for Flash Speculative Decoding},
  author    = {Chen, Jian and Liang, Yesheng and Liu, Zhijian},
  booktitle = {International Conference on Machine Learning (ICML)},
  year      = {2026}
}

@article{cheng2026dspark,
  title   = {{DSpark}: Confidence-Scheduled Speculative Decoding with Semi-Autoregressive Generation},
  author  = {Cheng, Xin and Yu, Xingkai and Shao, Chenze and Li, Jiashi and others},
  journal = {arXiv preprint arXiv:2607.05147},
  year    = {2026}
}

@inproceedings{giannou2023looped,
  title        = {Looped Transformers as Programmable Computers},
  author       = {Giannou, Angeliki and Rajput, Shashank and Sohn, Jy-yong and Lee, Kangwook and Lee, Jason D. and Papailiopoulos, Dimitris},
  booktitle    = {International Conference on Machine Learning (ICML)},
  year         = {2023},
  organization = {PMLR}
}

@inproceedings{dehghani2019universal,
  title     = {Universal Transformers},
  author    = {Dehghani, Mostafa and Gouws, Stephan and Vinyals, Oriol and Uszkoreit, Jakob and Kaiser, {\L}ukasz},
  booktitle = {International Conference on Learning Representations (ICLR)},
  year      = {2019}
}

@inproceedings{banino2021pondernet,
  title     = {{PonderNet}: Learning to Ponder},
  author    = {Banino, Andrea and Balaguer, Jan and Blundell, Charles},
  booktitle = {ICML Workshop on Automated Machine Learning (AutoML)},
  year      = {2021}
}

@inproceedings{schuster2022calm,
  title     = {Confident Adaptive Language Modeling},
  author    = {Schuster, Tal and Fisch, Adam and Gupta, Jai and Dehghani, Mostafa and Bahri, Dara and Tran, Vinh Q. and Tay, Yi and Metzler, Donald},
  booktitle = {Advances in Neural Information Processing Systems (NeurIPS)},
  year      = {2022}
}

@article{zhu2025ouro,
  title   = {Scaling Latent Reasoning via Looped Language Models},
  author  = {Zhu, Rui-Jie and Wang, Zixuan and Hua, Kai and Zhang, Tianyu and Li, Ziniu and Que, Haoran and Wei, Boyi and Wen, Zixin and Yin, Fan and others},
  journal = {arXiv preprint arXiv:2510.25741},
  year    = {2025}
}

@article{zou2026vsd,
  title   = {Variational Speculative Decoding: Rethinking Draft Training from Token Likelihood to Sequence Acceptance},
  author  = {Zou, Xiandong and Li, Jianshu and Huang, Jing and Zhou, Pan},
  journal = {arXiv preprint arXiv:2602.05774},
  year    = {2026}
}

@inproceedings{chen2024sequoia,
  title     = {Sequoia: Scalable, Robust, and Hardware-aware Speculative Decoding},
  author    = {Chen, Zhuoming and May, Avner and Svirschevski, Ruslan and Huang, Yuhsun and Ryabinin, Max and Jia, Zhihao and Chen, Beidi},
  booktitle = {Advances in Neural Information Processing Systems (NeurIPS)},
  year      = {2024}
}

@article{cheng2024redrafter,
  title   = {Recurrent Drafter for Fast Speculative Decoding in Large Language Models},
  author  = {Cheng, Yunfei and Zhang, Aonan and Zhang, Xuanyu and Wang, Chong and Wang, Yi},
  journal = {arXiv preprint arXiv:2403.09919},
  year    = {2024}
}

@inproceedings{elhoushi2024layerskip,
  title     = {{LayerSkip}: Enabling Early Exit Inference and Self-Speculative Decoding},
  author    = {Elhoushi, Mostafa and Shrivastava, Akshat and Liskovich, Diana and Hosmer, Basil and Wasti, Bram and Lai, Liangzhen and Mahmoud, Anas and Acun, Bilge and Agarwal, Saurabh and Roman, Ahmed and Aly, Ahmed and Chen, Beidi and Wu, Carole-Jean},
  booktitle = {Proceedings of the Annual Meeting of the Association for Computational Linguistics (ACL)},
  year      = {2024}
}

@inproceedings{liu2024kangaroo,
  title     = {Kangaroo: Lossless Self-Speculative Decoding via Double Early Exiting},
  author    = {Liu, Fangcheng and Tang, Yehui and Liu, Zhenhua and Ni, Yunsheng and Han, Kai and Wang, Yunhe},
  booktitle = {Advances in Neural Information Processing Systems (NeurIPS)},
  year      = {2024}
}

@inproceedings{svirschevski2024specexec,
  title     = {{SpecExec}: Massively Parallel Speculative Decoding for Interactive {LLM} Inference on Consumer Devices},
  author    = {Svirschevski, Ruslan and May, Avner and Chen, Zhuoming and Chen, Beidi and Jia, Zhihao and Ryabinin, Max},
  booktitle = {Advances in Neural Information Processing Systems (NeurIPS)},
  year      = {2024}
}

@inproceedings{zhang2024draftverify,
  title     = {Draft \& Verify: Lossless Large Language Model Acceleration via Self-Speculative Decoding},
  author    = {Zhang, Jun and Wang, Jue and Li, Huan and Shou, Lidan and Chen, Ke and Chen, Gang and Mehrotra, Sharad},
  booktitle = {Proceedings of the Annual Meeting of the Association for Computational Linguistics (ACL)},
  year      = {2024}
}

@inproceedings{zhang2025hass,
  title     = {Learning Harmonized Representations for Speculative Sampling},
  author    = {Zhang, Lefan and Wang, Xiaodan and Huang, Yanhua and Xu, Ruiwen},
  booktitle = {International Conference on Learning Representations (ICLR)},
  year      = {2025}
}

\appendix

\section{Details on Parallel Reconsideration}
\label{app:method}

\subsection{Reconsidering in parallel}
\label{app:method_parallel}

\begin{figure*}[t]
    \centering
    \includegraphics[width=\textwidth]{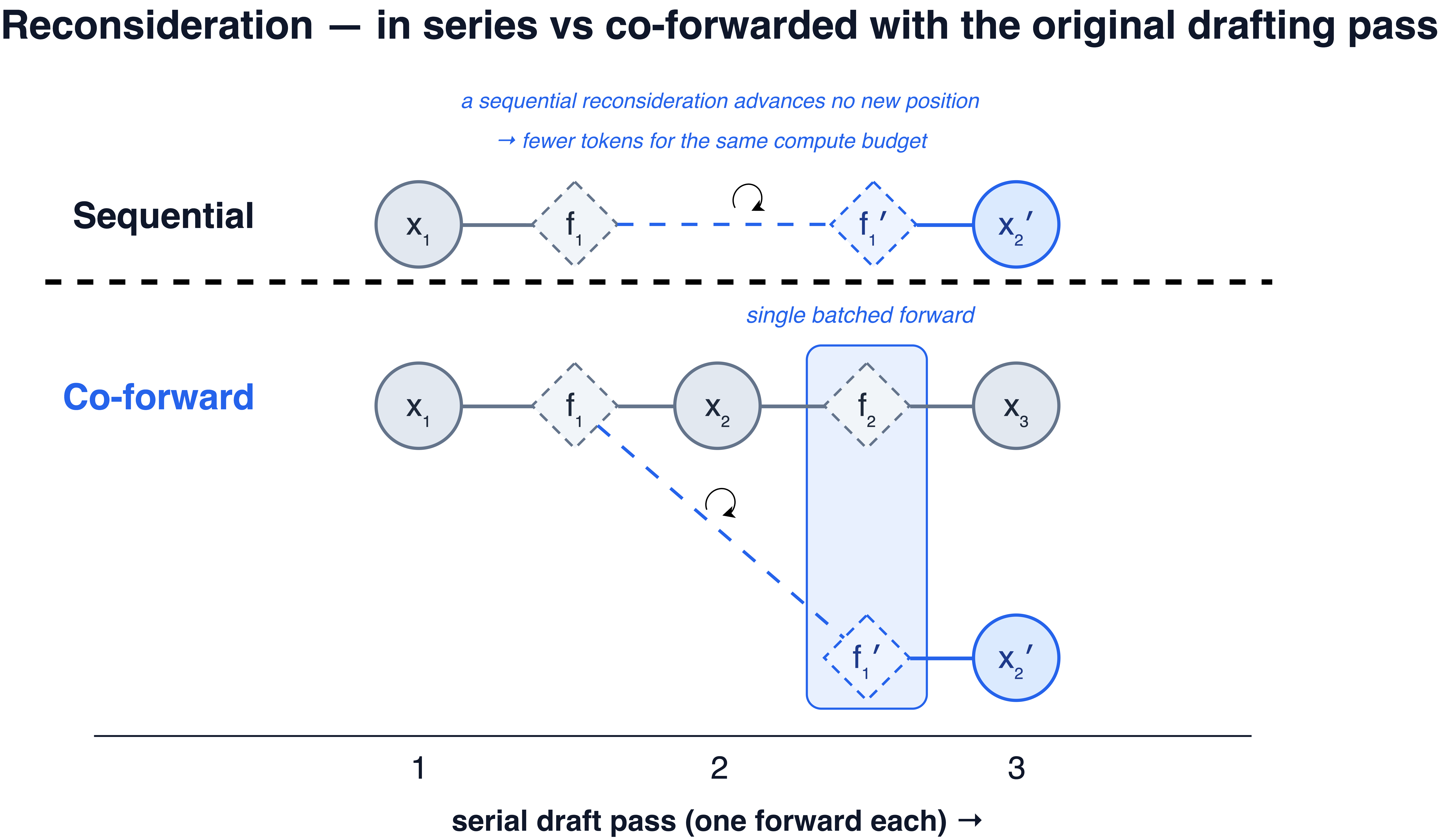}
    \caption{\textbf{Reconsidering in parallel.} Diamonds are features produced by a forward
        pass of the shared block $M$ and circles are the tokens sampled from them; gray marks
        ordinary one-pass drafting (feature $f_i$, token $x_i$) and blue marks reconsideration
        (deepened feature $f_i'$, reconsidered token $x_i'$).
        \emph{Top (sequential):} running reconsideration in series spends a forward of $M$ that
        advances no new position, so within a fixed draft budget the drafter reaches fewer tokens.
        \emph{Bottom (co-forward):} reconsideration of a position depends only on its one-pass
        feature, which is already available, so it is co-forwarded with the ordinary draft step
        in a single batched pass of $M$. The one-pass chain still reaches $x_3$ and the
        reconsidered token $x_2'$ is obtained for negligible additional latency.}
    \label{fig:coforward}
\end{figure*}

Reconsideration adds forward passes, and naively those passes are sequential.
Running the second pass in series stalls the draft at $p$ until $f_p^{1}$ is
computed, so a draft of $\gamma$ positions with $N$ reconsidered ones takes
$(\gamma+N)\,t_{\mathrm{step}}$ instead of $\gamma\,t_{\mathrm{step}}$
(Eq.~\eqref{eq:tdraft}, Figure~\ref{fig:coforward}, top). Under a fixed drafting-time
budget the drafter reaches $N$ fewer positions, shortening the original total number of
guesses.

The stall, however, is an artifact of ordering, not of dependency. Writing the draft
step and its reconsideration side by side (Eq.~\eqref{eq:coforward}),
both take the same feature $f_p^{0}$ as input, and their outputs go to separate
places: $f_{p+1}^{0}$ advances the ongoing draft to $p{+}1$, whereas $f_p^{1}$ only
re-samples $\tilde t_{p+1}$ and later seeds the reconsidered branch's own continuation,
which the ongoing draft never reads. Concretely, the reconsideration row keeps position
index $p$ and attends to the committed prefix $\kappa_{<p}$ and its own key, not to
$\mathrm{kv}(f_p^{0})$; its key/value replaces slot $p$ on the reconsidered branch only. The reconsideration is therefore independent of
the draft step that advances past it, so the two can be batched together into a
single forward of $M$ --- \emph{co-forwarding} rather than serializing
(Figure~\ref{fig:coforward}, bottom).

At the small batch sizes of latency-oriented decoding a forward of $M$ is
memory-bandwidth bound: its latency is set by streaming the drafter's weights from
HBM, and the extra arithmetic from enlarging the batch is hidden beneath that load. A
co-forwarded pass therefore costs $t_{\mathrm{step}}+\epsilon$ for a negligible
$\epsilon$, so a draft of $\gamma$ positions costs
$\gamma\,(t_{\mathrm{step}}+\epsilon) \approx \gamma\,t_{\mathrm{step}}$, against
$(\gamma+N)\,t_{\mathrm{step}}$ for the serial version. The drafter reaches exactly as
many positions as it would without reconsideration; we detail the batched execution in
Section~\ref{sec:serving}. Measured end to end, carrying the reconsideration rows costs
$1.015$--$1.022\times$ the per-cycle time of the one-pass drafter, i.e., roughly $2\%$ of
cycle time (Appendix~\ref{app:coforward} for details).

The result is an \emph{interleaved} drafting strategy: the ordinary steps form the main
branches of one-pass guesses, and each gated position might add a reconsidered candidate
$\tilde t_{p+1}$ (Eq.~\eqref{eq:reconsider}) beside its one-pass sibling.
Verification scores both, so at that position the reconsidered token takes nothing from
its one-pass sibling: it is an extra candidate rather than a substitute, and it can
supply an accepted token only where the one-pass guess would have been rejected. The
two share the verification budget, however, and Section~\ref{sec:serving} is where
they are ranked against each other for it.

\section{Details on Training Parallel Reconsideration}
\label{app:training}

This appendix gives a more detailed discussion on the training recipe of Parallel
Reconsideration. Section~\ref{sec:training} states the objective of
Eq.~\eqref{eq:fullloss} and the mixed cache of Eq.~\eqref{eq:haltkv}; we expand here on
the two failure modes the recipe is designed against
(Appendix~\ref{app:train_chaina}), on what a reconsidered feature is actually supervised
to do (Appendix~\ref{app:train_recloss}), on the rollouts the losses are computed over,
and on why the balance term regularizes the gate's marginal usage rather than its
per-position halt distribution (Appendix~\ref{app:train_objective}).

\subsection{Why the one-pass head cannot be scaffolding}
\label{app:train_chaina}

Loop transformers, which reuse a layer across depths much as we do, read out only at a
single depth and are free to supervise the intermediate depths purely as scaffolding for
that one output~(Appendix~\ref{app:related_work}). Parallel Reconsideration cannot do
this, because at serving \emph{both} depths are read out and committed to the tree: the
one-pass head guesses the token at most positions and the reconsidered head revises it at
the uncertain ones. The difficulty is that the two share the feature $f_p^{0}$. On one
hand, $f_p^{0}$ is the feature every later draft step conditions on, so it must retain
enough of the target distribution to keep proposing acceptable tokens down the whole
draft; on the other hand, reconsideration deepens $f_p^{0}$ into $f_p^{1}$, and for the
reconsidered token to differ from the one-pass token the objective must let $f_p^{1}$
diverge from $f_p^{0}$ rather than reproduce it.

This is what makes the two failure modes of Section~\ref{sec:training} opposite rather
than two versions of the same error. Supervising only the reconsidered head leaves the
one-pass head untrained, so the draft it proposes degrades and shortens, and since a
reconsidered candidate is only reached once the tokens before it are accepted, the
reconsidered branches are left with shorter and shorter prefixes to extend. Supervising
both to match the target with equal weight instead removes any incentive for the two
features to differ, and gradient descent finds the trivial optimum
$f_p^{1}\!\approx\!f_p^{0}$. Appendix~\ref{app:train_ablation} discusses both in more
detail.

The recipe therefore holds the one-pass draft to ordinary EAGLE quality with an
unconditional loss,
\begin{equation}
    \mathcal{L}^{\mathrm{one}}_p = \mathrm{CE}\!\left(\mathrm{head}(f_p^{0}),\, t_{p+1}^{\star}\right),
    \label{eq:chaina}
\end{equation}
against the teacher token $t_{p+1}^{\star}$, which appears as the unweighted first term
of Eq.~\eqref{eq:fullloss}. Supervising it at every position frees reconsideration to
specialize on the positions the one-pass guess hesitates over, without eroding the ones
it already resolves.

\subsection{What a reconsidered feature is supervised to do}
\label{app:train_recloss}

Each reconsideration pass reapplies $M$ to the previous depth's feature through the
self-query, $f_p^{r} = M(\mathrm{fc}(f_p^{r-1}, Q(f_p^{r-1})))$
(Eq.~\eqref{eq:reconsider}), so depth $r$ is a strictly deeper reading of the same
position $p$. Such a feature does two jobs at serving, and the two are supervised
differently. \emph{Every} depth is read out to propose a candidate token, so we supervise
the head at all depths. Only \emph{one} depth, the halt depth $d_p$ where the gate stops,
has its key/value written to the cache and drafts the next position, so the continuation
is supervised at that depth alone. Which depth halts is therefore central: it decides
what the deeper positions read and what the recurrence is trained on. We sample it,
$d_p \sim q$ (Eq.~\eqref{eq:gate}), by walking the depths and halting at the first pass
the gate declines, the \emph{same} way in training and at serving.

\paragraph{The mixed cache, concretely.}
Section~\ref{sec:training} introduces the mixed cache of Eq.~\eqref{eq:haltkv}. Training
naively on a fixed one-pass cache, reading $f_j^{0}$ at every ancestor, would leave a
gap: that cache never appears at serving, where each ancestor commits at its own halt
depth. For a position $5$ whose ancestors halt at depths $d_{1:4}=(0,1,0,2)$,
\begin{align}
    \text{one-pass cache:}\quad
     & \big(\mathrm{kv}(f_1^{0}),\,\mathrm{kv}(f_2^{0}),\,\mathrm{kv}(f_3^{0}),\,\mathrm{kv}(f_4^{0})\big),
    \nonumber                                                                                               \\
    \text{mixed cache } \kappa_{<5}:\quad
     & \big(\mathrm{kv}(f_1^{0}),\,\mathrm{kv}(f_2^{1}),\,\mathrm{kv}(f_3^{0}),\,\mathrm{kv}(f_4^{2})\big),
    \label{eq:mixedexample}
\end{align}
the two agree at the confident ancestors $1,3$ (halt depth $0$) but differ at the
uncertain $2,4$, where the mixed cache carries the reconsidered key/value the drafter
will actually commit. Sampling $d_j \sim q$ deepens the ancestors of a single rollout
faithfully under the halting distribution, while a served branch under a budget deepens
few; both are prefixes the same per-position commit rule produces, so a reconsidered
branch never reads one the drafter was not trained for.

\paragraph{On the halt depths seen at training and at serving.}
The depths $q$ favors at training are not the depths the drafter meets at serving.
Training samples $d_p \sim q$ at every position, so any position may deepen; at serving
the reconsideration budget $m$ admits only a few eligible nodes per step, so most
positions are committed from depth $0$ however willing $q$ is to go deeper. The
weighting does not turn this into a train--serve gap, because it reallocates supervision
rather than withholding it. Every depth is read out and scored at \emph{every} position,
and only the size of its gradient varies with $q(r)$, so no depth is trained solely on
the positions $q$ favors and each stays accurate wherever it is eventually read. A
position served from depth $0$ under a tight budget is one the one-pass term trains
unconditionally; a position the budget does deepen is one the gated terms have trained at
that depth. The same holds for the cache those losses are computed over, by the argument
above.

\paragraph{Two on-policy rollouts.}
EAGLE-3~\citep{li2025eagle3} trains by unrolling an on-policy draft chain from every
position of the training sequence: starting from a position's target feature, the
drafter speculates several steps forward on its own predictions, and each drafted
position is scored against the teacher token. We unroll two such chains from every
starting position. The \emph{one-pass chain} is EAGLE-3's own, advancing from each
one-pass feature to the next, $f_p^{0}\!\to\!f_{p+1}^{0}$. The \emph{reconsideration
    chain} branches off at the starting position, then runs on its own features: at each
of its positions it deepens the current feature to the halt feature $f_p^{\,d_p}$ and
drafts the next position from it,
\begin{equation}
    f_{p+1}^{0} = M\!\left(\mathrm{fc}\big(f_p^{\,d_p},\, e(t_{p+1})\big)
    \;\middle|\; \kappa_{<p+1}\right),
    \label{eq:carry}
\end{equation}
reading the mixed cache assembled above and reconsidering afresh from there.

Both chains are scored, position by position, by the head losses of
Eq.~\eqref{eq:fullloss}, so each learns its own predictions on its own rollout. They
share parameters through $M$ but couple in the backward pass only at the branch point
$f_p^{0}$, which feeds both the one-pass step and the reconsideration chain. Its gradient
is therefore the sum of two signals: the one-pass head loss at $f_p^{0}$ and the losses
of the entire reconsideration rollout it seeds. Thus $f_p^{0}$ is optimized jointly to
minimize the one-pass token error and to encode a state from which reconsideration can
recover the tokens that error misses. The continuation of Eq.~\eqref{eq:carry} needs no
term of its own: keeping each chain accurate at every position already trains its
features to draft the next one.

\subsection{Which entropy the balance term rewards}
\label{app:train_objective}

The balance term of Eq.~\eqref{eq:fullloss} is not the obvious entropy. The halt
distribution induces two: the per-position entropy averaged over positions,
$\mathbb{E}_p[H(q)]$, and the entropy of the marginal usage $\bar q = \mathbb{E}_p[q]$.
We reward the marginal $H(\bar q)$, and the decomposition of
Proposition~\ref{prop:entropy} (Appendix~\ref{app:entropy}) makes the choice precise.

The point is what each term does \emph{not} do. Rewarding per-position entropy
$\mathbb{E}_p[H(q)]$ would push every position to split its mass across depths and
actively destroy the specialization we are trying to build. Rewarding the marginal
$H(\bar q)$ leaves that specialization alone: it charges only for letting a depth fall
out of use, and is neutral as to whether the surviving mass is realized by confident
per-position decisions or by undecided ones. The specialization itself comes from the
head term of Eq.~\eqref{eq:fullloss}, which trains each depth on the tokens routed to it;
the balance term's job is only to keep a depth from going dark before that term can train
it. It is also symmetric in the depths: a gate that always halts at depth $0$ and one
that always runs to depth $R$ both give $H(\bar q)=0$, so it resists collapse onto any
single depth rather than pulling toward deeper ones~\citep{zhu2025ouro}.

\paragraph{No price on depth.}
No term of Eq.~\eqref{eq:fullloss} charges for reconsidering. With the losses detached,
$\sum_{r} q(r)\,\mathrm{sg}[\ell_p^{r}]$ is minimized by routing each position to its
lowest-loss depth, and nothing opposes going deeper. The objective therefore asks the
gate only whether a further pass improves the token, which is what
Section~\ref{sec:method_gate} wants it to judge; how many passes the drafter can afford
is set where the resource binds, by the depth cap $R$ and the $m$ reconsideration rows a
forward carries (Section~\ref{sec:serving}).

\section{Details on Serving a Reconsidered Tree}
\label{app:serving}

\begin{algorithm}[t!]
    \DontPrintSemicolon
    \SetKwInOut{Input}{input}
    \SetKwInOut{Output}{output}
    \Input{budget $B$; grow width $B_w$; offer width $k$; reconsideration budget $m$;
        reconsideration cap $R$; draft steps $\gamma$}
    \Output{draft tree $\mathcal{T}$, $|\mathcal{T}| \le B$}
    \BlankLine
    $\mathcal{T} \leftarrow \{\text{root}\}$;\quad
    $\mathcal{U} \leftarrow \{\text{root}\}$;\quad
    $\mathcal{C} \leftarrow \emptyset$
    \tcp*{tree; un-forwarded nodes; reconsideration pool}
    \For{$\mathrm{step} = 1,\dots,\gamma$}{
        $\mathcal{A} \leftarrow \text{top-}B_w \text{ of } \mathcal{U} \text{ by } s(\cdot)$;\quad
        $\mathcal{U} \leftarrow \mathcal{U} \setminus \mathcal{A}$
        $\mathcal{S} \leftarrow \text{top-}m \text{ of } \mathcal{C} \text{ by } \mathrm{bid}(\cdot)$
        \tcp*{Eq.~\eqref{eq:bid}; the $m$ most promising of the pool}
        \BlankLine
        $\big\{f_u^{0}\big\}_{u \in \mathcal{A}}
            \;\cup\;
            \big\{f_v^{\,r_v+1}\big\}_{v \in \mathcal{S}}
            \;\leftarrow\; M\big(\mathcal{A} \cup \mathcal{S}\big)$
        \tcp*{one level deeper per selected node}
        \BlankLine
        \ForEach{$v \in \mathcal{S}$}{
            halt with probability $\lambda_{r_v} = \sigma\big(\mathrm{gate}(f_v^{\,r_v})\big)$, Eq.~\eqref{eq:gate}
            \tcp*{sampled, as in training}
            \eIf{halted or $r_v = R$}{$d_v \leftarrow r_v$; remove $v$ from $\mathcal{C}$}{$r_v \leftarrow r_v + 1$; keep $v$ in $\mathcal{C}$ for a later step}
        }
        \BlankLine
        \tcc{two chains: a reconsidered node adds to what it already offered}
        \ForEach{$u \in \mathcal{A}$}{
            offer the top $k$ tokens of $\pi_{\mathrm{one}}(\cdot \mid u)$ as children of $u$,
            scoring each by $s(c) = s(u) + \log \pi_{\mathrm{one}}(c \mid u)$\;
        }
        \ForEach{$v$ halted this step}{
            \tcc{$v$ now prices its edges by $\pi_{\mathrm{mix}}$: reprice the children it
                offered when it was advanced, then admit the ones the halt distribution adds}
            \ForEach{existing child $c$ of $v$}{
                $\delta_c \leftarrow \log \pi_{\mathrm{mix}}(c \mid v) - \log \pi_{\mathrm{one}}(c \mid v)$,
                Eq.~\eqref{eq:refprice}\;
                add $\delta_c$ to $s(c)$ and to every node beneath $c$
                \tcp*{$s(\cdot)$ sums along the path}
            }
            \ForEach{$t$ with $q(d_v)\big(\pi_{d_v}(t \mid v) - \pi_{\mathrm{one}}(t \mid v)\big) > 0$
                and $t$ not already a child of $v$}{
                offer $t$ as a child $c$ of $v$ with
                $s(c) = s(v) + \log \pi_{\mathrm{mix}}(t \mid v)$\;
            }
        }
        \BlankLine
        $\mathcal{U} \leftarrow \mathcal{U} \cup \{\text{children offered this step}\}$;\quad
        $\mathcal{T} \leftarrow \mathcal{T} \cup \{\text{children offered this step}\}$\;
        \ForEach{$u \in \mathcal{A}$}{
            $r_u \leftarrow 1$; add $u$ to $\mathcal{C}$ with probability
            $1 - \lambda_0 = 1 - \sigma(\mathrm{gate}(f_u^{0}))$
            \tcp*{gate eligibility at depth $0$, free from $f_u^{0}$}
        }
    }
    \Return $\text{top-}B \text{ of } \mathcal{T} \text{ by } s(\cdot)$
    \tcp*{one global cut, one-pass and reconsidered alike}
    \caption{One draft round of Parallel Reconsideration. A forwarded node joins the
        reconsideration pool $\mathcal{C}$ only if the gate declines to halt it at depth
        $0$. The bid then selects the $m$ eligible nodes whose reconsideration is most
        worth the pass, and the gate sets how deep each goes. With $\mathcal{C}$ held
        empty the procedure is the EAGLE-3.1 loop.}
    \label{alg:serving}
\end{algorithm}

This appendix gives a more detailed discussion on the serving algorithm of Parallel
Reconsideration, summarized in Algorithm~\ref{alg:serving}. We expand on how the
reconsideration pool is maintained across draft steps
(Appendix~\ref{app:serving_pool}), on what the repricing rule of
Eq.~\eqref{eq:refprice} does to a node's existing children as well as its new ones
(Appendix~\ref{app:serving_price}), on why the round degenerates exactly to
EAGLE-3.1 when no node is eligible (Appendix~\ref{app:serving_degenerate}), and on why
verification, and hence losslessness, is unaffected (Appendix~\ref{app:serving_lossless}).

\subsection{Maintaining the reconsideration pool}
\label{app:serving_pool}

The gate is consulted once per level, always on the feature the last forward produced, so
no evaluation of it costs a pass of its own. At level $0$ this is what admits a node to
the pool: the step that forwards $u$ reads
$\lambda_0 = \sigma(\mathrm{gate}(f_u^{0}))$ off the one-pass feature it has just
produced, and $u$ enters the pool with probability $1-\lambda_0$, sampled as in training
(Section~\ref{sec:method_gate}). Eligibility is therefore settled before any
reconsideration pass is spent, which is what lets the pool be filtered for free.

At level $r \ge 1$ the same test decides how much further a node already in the pool
goes. Each step the bid selects the top $m$ of the pool (Eq.~\eqref{eq:bid}); those nodes
are carried one level deeper in the same batched forward, and the gate then reads
$\lambda_{r_v}$ off the feature $f_v^{\,r_v}$ that forward produced. A node it halts
leaves the pool with its halt depth $d_v$ fixed and reprices its edges; a node it
declines to halt has $r_v$ incremented and remains for a later step, so a single position
may be reconsidered across several steps up to the cap $R$, at which it halts regardless.
Two consequences follow.
First, the bid is re-evaluated every step over a changing pool, so a node that loses the
competition early may still be selected once shallower competitors have halted. Second,
the $m$ rows a step carries are not tied to any one position: they are spent on whichever
$m$ pooled nodes currently bid highest, which is what makes $m$ a per-step budget rather
than a per-position one.

\subsection{What the repricing rule covers}
\label{app:serving_price}

Once $v$ halts at $d_v$, Eq.~\eqref{eq:refprice} prices \emph{every} edge out of it, not
only the token the reconsidered feature newly proposes. The children its one-pass feature
offered when it was advanced are repriced by the same rule, and because the path score
$s(\cdot)$ is a sum of edge log-probabilities, that correction carries into the subtree those children have
already grown: in the algorithm this is the increment
$\delta_c = \log \pi_{\mathrm{mix}}(c \mid v) - \log \pi_{\mathrm{one}}(c \mid v)$ added to
$s(c)$ and to every node beneath $c$. A node the bid never selects has no $\pi_{d_v}$, so
Eq.~\eqref{eq:refprice} loses its second term and returns $\pi_{\mathrm{one}}$, leaving its
edges exactly as the one-pass draft priced them.

This is where reconsideration does more than widen the tree. On a token $v$ has not
proposed, a positive correction admits a new candidate. On one it already holds, the
correction moves that child instead, up where the deeper pass agrees with it and down
where it does not, against its siblings and against the rest of the tree at the cut.

\subsection{Degenerating to the one-pass baseline}
\label{app:serving_degenerate}

The round reduces to EAGLE-3.1's exactly when the pool is empty. With
$\mathcal{C} = \emptyset$ no node is ever selected by the bid, no forward carries
reconsideration rows, every node keeps $\pi_{\mathrm{one}}$ by the degenerate case of
Eq.~\eqref{eq:refprice}, and the global cut ranks a tree grown entirely by one-pass
steps. This is the sense in which reconsideration is additive at serving: the gate
controls whether the second mechanism runs at all, and where it declines, the drafter
behaves as the baseline it was built on.

\subsection{Verification is unchanged, and the output is lossless}
\label{app:serving_lossless}

Reconsideration changes \emph{which} tokens enter the draft tree, but not how the tree
is verified. The verifier receives only the tree's tokens and its topology. It never
receives a draft distribution: neither $\pi_{\mathrm{one}}$, nor $\pi_{d_v}$, nor
$\pi_{\mathrm{mix}}$, nor any path score $s(\cdot)$. These quantities, together with the
sampled halting decisions, are used only on the drafter side, to decide what to propose
and what survives the global cut. We therefore serve Parallel Reconsideration through
SGLang's tree verification for EAGLE, unmodified, the same path used for the EAGLE-3.1
and HASS baselines. At temperature $0$ it accepts the longest path that matches the
target's greedy tokens. At temperature $T>0$ it applies target-only multi-candidate
rejection sampling, which treats every drafted child as a deterministic proposal.

\paragraph{Losslessness.}
Fix a verified node with prefix $x$, and let $\pi = \pi_{\mathrm{tgt}}(\cdot \mid x)$ be
the target's distribution at temperature $T$. Let $c_1, \dots, c_k$ be the node's
children in the order the verifier visits them. The verifier sets $r_0 = \pi$. For
$i = 1, \dots, k$, it accepts $c_i$ with probability $r_{i-1}(c_i)$. On rejection it
sets $r_i = r_{i-1}(\cdot \mid \cdot \neq c_i)$, that is, $r_{i-1}$ with $c_i$ removed
and the rest renormalized. If every child is rejected, it samples the emitted token from
$r_k$. Sampling from $r_{i-1}$ is the same as emitting $c_i$ with probability
$r_{i-1}(c_i)$ and otherwise sampling from $r_{i-1}$ conditioned on $\neq c_i$, which is
$r_i$. By induction on $i$, the emitted token is therefore distributed exactly as
$r_0 = \pi$, for \emph{every} choice of $c_1, \dots, c_k$, duplicates included (a
repeated child has zero residual mass and is never accepted twice). The only requirement
is that the children are fixed before the verifier draws its coins, which holds because
drafting finishes before verification begins. Applying the argument node by node along
the accepted path shows that each emitted token is an exact sample from the target given
its prefix. The output sequence is thus distributed as autoregressive sampling from the
target.

Nothing in this argument refers to how the candidates were produced. Whether a child
was drawn from $\pi_{\mathrm{one}}$ or from $\mathrm{head}(f_v^{d_v})$, whether its edge
was repriced by Eq.~\eqref{eq:refprice}, and whether the gate halted stochastically,
all of this changes only the candidate set, and hence \emph{how many} tokens are
accepted per round, never \emph{which distribution} they are drawn from. The same holds
for EAGLE-3.1, whose tree is also chosen by drafter-side scores that the verifier never
sees. Reconsideration therefore adds nothing to verification that a standard
speculative decoder does not already rely on, and the temperature-$1$ results in
Table~\ref{tab:main_results} are lossless in the same sense as the baseline's.

\section{Details on Experiments}
\label{app:experiments}

This appendix gives the full experimental detail behind Section~\ref{sec:experiments}. We
state the setup in full (Appendix~\ref{app:exp_setup}), then report four further studies:
whether spending the same compute on a wider draft
reproduces the gain (Appendix~\ref{app:exp_ablation}), whether the gain holds across
tree configurations (Appendix~\ref{app:exp_budget}) and on top of a different base
drafter (Appendix~\ref{app:exp_hass}), and which of the drafter's
failures reconsideration actually recovers, returning to the diagnosis of
Section~\ref{sec:moti_diagnosis} (Appendix~\ref{app:exp_recovery}).

\subsection{Experimental setup}
\label{app:exp_setup}

\paragraph{Models and baseline.}
We evaluate on three target models, \textbf{Llama-3.1-8B-Instruct},
\textbf{Qwen3-8B}, and \textbf{Gemma-3-12B}, and compare \textbf{Parallel
    Reconsideration} against \textbf{EAGLE-3.1}, the state-of-the-art feature-level
speculative decoder. We use EAGLE-3.1 to denote the current form of the EAGLE-3
drafter~\citep{li2025eagle3} with the normalization fixes of
\citet{eldenk2026attention}. Both drafters use a single decoder
layer and share an identical target model, and both are trained on
UltraChat~\citep{ding2023ultrachat} under EAGLE-3's released recipe, unchanged
(Appendix~\ref{app:impl}). Parallel Reconsideration takes at most two reconsideration
steps per position (depth cap $R=2$) on every model.

\paragraph{Implementation.}
We train both methods in SpecForge~\citep{li2026specforge}, publicly available training recipe of
EAGLE-3.1, and add reconsideration as a second head and gate within the same
trainer. Both are then served
through SGLang~\citep{zheng2024sglang}, and every acceptance length and speedup we
report is measured on SGLang serving engine, faithfully reporting realistic scenarios. As an external check on the baseline, our EAGLE-3.1 acceptance lengths agree with
those reported independently for the same models and benchmarks
by DFlash~\citep{chen2026dflash}.

\paragraph{Benchmarks and metrics.}
We evaluate on seven benchmarks in three categories: \emph{math}
(\textbf{GSM8K}~\citep{cobbe2021gsm8k}, \textbf{MATH-500}~\citep{lightman2024math500},
\textbf{AIME25}~\citep{maa2025aime}), \emph{code}
(\textbf{HumanEval}~\citep{chen2021humaneval}, \textbf{MBPP}~\citep{austin2021mbpp},
\textbf{LiveCodeBench}~\citep{jain2025livecodebench}), and \emph{chat}
(\textbf{MT-Bench}~\citep{zheng2023mtbench}). Our primary
metric is the \emph{acceptance length} $\tau$, the mean number of tokens the target
accepts per verification round, which is the drafter's contribution to end-to-end speedup
at fixed target and verifier cost (Eq.~\eqref{eq:costmodel}); we also report the
wall-clock speedup over autoregressive decoding, measured on one H200. We report both greedy decoding (temperature $0$) and
sampling at temperature $1$, with max generation length of $2048$ tokens. Benchmark sizes
and sources are in Appendix~\ref{app:impl}.

\paragraph{Budget configuration.}
Both drafters serve a dynamic draft tree bounded by a total budget $B$ (the number of
draft nodes verified per round), grown over $\gamma$ draft steps with offer width $k$
and grow width $B_w$ (Section~\ref{sec:serving}). We hold all of these equal across the
two drafters: $B \in \{24, 48\}$, shown in parentheses in Table~\ref{tab:main_results},
with $k=6$, $B_w=6$, and $\gamma=8$. Parallel Reconsideration additionally batches at
most $m=3$ reconsidered nodes per draft step, which share the co-forwarded batch;
Appendix~\ref{app:exp_ablation} grants the baseline the same enlarged batch and spends it
on one-pass breadth instead.

\subsection{Reconsidering against drafting wider}
\label{app:exp_ablation}

Holding $B$, $k$, $B_w$ and $\gamma$ equal leaves one asymmetry: a draft step of Parallel
Reconsideration batches $B_w + m$ nodes where EAGLE-3.1 batches $B_w$. If enlarging the
batch is nearly free, then EAGLE-3.1 can enlarge its own by $m$ and spend the additional
nodes on one-pass breadth, and if that recovers the same acceptance then the gain belongs
to the wider batch rather than to reconsideration.

We therefore grant the baseline exactly that, in two ways. The first control draws $m$
nodes from the previous step's batch and, for each such node $v$, forwards with an ordinary
draft step its best one-pass child not yet forwarded, i.e., a sibling of the child already
advanced. These rows take new inputs and so compute new features, growing a second branch
beneath $v$; the only difference is tree growing against reconsideration. Three properties
are held fixed. The \emph{source set} is the same, the previous step's batch rather than
the current frontier, since reconsideration cannot reach further either. The
\emph{final pruning} is the same, in that the children they grow compete for the same places in the tree on
the same score. And the \emph{batch is the same size}, $B_w = 6$
advanced nodes plus $m = 3$ more.

The second control drops the first of those constraints and simply widens the grow width
to $B_w = 9$. All nine nodes now extend the current frontier, which is more drafting
breadth than Parallel Reconsideration has: our own $m$ nodes are confined to positions
already forwarded, whereas these are free to advance the best of the live frontier.

As a sanity check in the other direction, we also run Parallel Reconsideration with
$m=0$: the same drafter and serving path, but with no node ever reconsidered, so each
step batches the same $B_w=6$ nodes as EAGLE-3.1.

\begin{table}[t]
    \centering
    \caption{\textbf{Reconsidering the additional nodes against drafting wider with them.}
        Gemma-3-12B at $B=48$, temperature $1$, reporting $\tau$ / speedup. The two
        EAGLE-3.1 controls batch $9$ nodes per draft step: the first takes $3$ nodes
        of the previous step's batch, matching where reconsideration draws from, and
        forwards each one's best not-yet-forwarded one-pass child (a new sibling branch); the second widens the grow width outright, so all $9$ nodes extend the
        current frontier, which is more drafting breadth than Parallel Reconsideration
        itself gets. Parallel Reconsideration with $m=0$ runs our full serving path
        with no node ever reconsidered and batches the same $6$ nodes as EAGLE-3.1;
        adding $3$ reconsidered nodes brings it to $9$. $\Delta$ is relative to EAGLE-3.1.}
    \label{tab:ablation_advance}
    \resizebox{\linewidth}{!}{
        \begin{tabular}{lccccccccc}
            \toprule
                                                 & \multicolumn{2}{c}{GSM8K} & \multicolumn{2}{c}{LiveCodeBench} & \multicolumn{2}{c}{MT-Bench} & \multicolumn{2}{c}{\textbf{Avg.}}                                                                                         \\
            \cmidrule(lr){2-3}\cmidrule(lr){4-5}\cmidrule(lr){6-7}\cmidrule(lr){8-9}
            \textbf{Drafter}                     & $\tau$                    & Speedup                           & $\tau$                       & Speedup                           & $\tau$            & Speedup               & $\tau$            & Speedup               \\
            \midrule
            EAGLE-3.1 ($B_w=6$)                  & 5.44                      & 2.81$\times$                      & 5.54                         & 3.13$\times$                      & 3.40              & 1.64$\times$          & 4.79              & 2.53$\times$          \\
            \quad + advance $3$ more             & 5.51                      & 2.83$\times$                      & 5.68                         & 3.17$\times$                      & 3.43              & 1.64$\times$          & 4.87              & 2.55$\times$          \\
            \quad\quad $\Delta$                  & $+1.3\%$                  & $+0.7\%$                          & $+2.5\%$                     & $+1.3\%$                          & $+0.9\%$          & $+0.0\%$              & $+1.7\%$          & $+0.8\%$              \\
            \quad widen to $B_w=9$               & 5.49                      & 2.82$\times$                      & 5.61                         & 3.15$\times$                      & 3.42              & 1.64$\times$          & 4.84              & 2.54$\times$          \\
            \quad\quad $\Delta$                  & $+0.9\%$                  & $+0.4\%$                          & $+1.3\%$                     & $+0.6\%$                          & $+0.6\%$          & $+0.0\%$              & $+1.0\%$          & $+0.4\%$              \\
            \midrule
            PR without reconsideration ($m=0$)   & 5.44                      & 2.81$\times$                      & 5.55                         & 3.13$\times$                      & 3.38              & 1.63$\times$          & 4.79              & 2.52$\times$          \\
            \quad\quad $\Delta$                  & $+0.0\%$                  & $+0.0\%$                          & $+0.2\%$                     & $+0.0\%$                          & $-0.6\%$          & $-0.6\%$              & $-0.1\%$          & $-0.4\%$              \\
            \textbf{\quad + reconsider $3$ (PR)} & \textbf{6.22}             & \textbf{3.19$\times$}             & \textbf{5.99}                & \textbf{3.37$\times$}             & \textbf{3.55}     & \textbf{1.70$\times$} & \textbf{5.25}     & \textbf{2.75$\times$} \\
            \quad\quad $\Delta$                  & $\mathbf{+14.3\%}$        & $\mathbf{+13.5\%}$                & $\mathbf{+8.1\%}$            & $\mathbf{+7.7\%}$                 & $\mathbf{+4.4\%}$ & $\mathbf{+3.7\%}$     & $\mathbf{+9.6\%}$ & $\mathbf{+8.7\%}$     \\
            \bottomrule
        \end{tabular}
    }
\end{table}

Without reconsideration, Parallel Reconsideration matches EAGLE-3.1 to within noise
($-0.1\%$ acceptance and $-0.4\%$ speedup on average), as
Appendix~\ref{app:serving_degenerate} predicts. The second head and gate therefore
neither help nor hurt on their own, and the serving path adds no measurable overhead
when it is idle. Neither control recovers the gain either. Advancing $3$ more lifts average acceptance by $1.7\%$ over
EAGLE-3.1 and widening to $B_w=9$ by $1.0\%$, against \textbf{9.6\%} for reconsidering
the same three, and speedup splits the same way ($+0.8\%$ and $+0.4\%$ against
$\mathbf{+8.7\%}$). That the freer of the two controls rules out breadth
as the explanation: the additional capacity pays only when it is spent on a second look
at an uncertain feature. Appendix~\ref{app:exp_recovery} examines which of the one-pass
drafter's misses reconsideration recovers.

\subsection{Sensitivity to the tree configuration}
\label{app:exp_budget}

Our main results fix one tree configuration ($B=48$, $B_w=6$, $k=6$) for both
drafters. To check that the gain is not an artifact of that choice, we vary the total
budget $B$, the grow width $B_w$ and the offer width $k$ one at a time around it,
again holding every knob equal across the two drafters (Table~\ref{tab:ablation_budget}).

\begin{table}[t]
    \centering
    \caption{\textbf{Sensitivity to the tree configuration.} Acceptance length $\tau$
        and speedup over autoregressive decoding for EAGLE-3.1 and Parallel
        Reconsideration (PR) under identical budgets, varying one of the total budget
        $B$, grow width $B_w$ and offer width $k$ at a time around our default
        ($B=48$, $B_w=6$, $k=6$, marked $\ast$ and repeated in each block); $\gamma=8$ and $m=3$
        throughout, on Qwen3-8B, GSM8K, temperature $0$. $\Delta\tau$ is PR minus EAGLE-3.1. Bold marks the best value per
        column.
    }
    \label{tab:ablation_budget}
    \resizebox{\linewidth}{!}{
        \begin{tabular}{ccc|ccc|cc}
            \toprule
                  &       &     & \multicolumn{3}{c|}{$\tau$} & \multicolumn{2}{c}{Speedup}                                                   \\
            \cmidrule(lr){4-6}\cmidrule(lr){7-8}
            $B$   & $B_w$ & $k$ & EAGLE-3.1                   & PR                          & $\Delta\tau$    & EAGLE-3.1            & PR                   \\
            \midrule
            \multicolumn{8}{l}{\emph{Total budget}}                                                                                                     \\
            24    & 6     & 6   & 4.39                        & 4.83                        & $\mathbf{+0.44}$ & 1.94$\times$         & 2.11$\times$         \\
            32    & 6     & 6   & 4.606                       & 4.869                       & $+0.263$        & 2.077$\times$        & 2.208$\times$        \\
            48$^\ast$ & 6     & 6   & 4.835                       & 5.127                       & $+0.292$        & 2.184$\times$        & \textbf{2.291$\times$} \\
            64    & 6     & 6   & \textbf{4.961}              & \textbf{5.222}              & $+0.261$        & \textbf{2.195$\times$} & 2.280$\times$        \\
            \midrule
            \multicolumn{8}{l}{\emph{Grow width}}                                                                                                       \\
            48    & 4     & 6   & 4.797                       & 5.069                       & $+0.272$        & 2.105$\times$        & 2.232$\times$        \\
            48$^\ast$ & 6     & 6   & 4.835                       & 5.127                       & $+0.292$        & 2.184$\times$        & \textbf{2.291$\times$} \\
            48    & 8     & 6   & 4.826                       & 5.126                       & $+0.300$        & 2.138$\times$        & 2.266$\times$        \\
            48    & 10    & 6   & 4.819                       & 5.124                       & $+0.304$        & 2.096$\times$      & 2.227$\times$        \\
            \midrule
            \multicolumn{8}{l}{\emph{Offer width}}                                                                                                      \\
            48    & 6     & 4   & 4.767                       & 5.061                       & $+0.294$        & 2.162$\times$        & 2.286$\times$        \\
            48$^\ast$ & 6     & 6   & 4.835                       & 5.127                       & $+0.292$        & 2.184$\times$        & \textbf{2.291$\times$} \\
            48    & 6     & 8   & 4.837                       & 5.124                       & $+0.287$        & 2.080$\times$        & 2.250$\times$        \\
            \bottomrule
        \end{tabular}
    }
\end{table}

The acceptance gain holds across all nine configurations: Parallel Reconsideration
adds $+0.26$ to $+0.30$ tokens per round over EAGLE-3.1, or $5.3$--$6.3\%$ in
relative terms, from $B=32$ to $B=64$ and whether the tree is narrow ($B_w=4$) or
wide ($B_w=10$), and the gain grows to $+0.44$ tokens ($+10.0\%$) at the smallest
budget ($B=24$). Reconsideration therefore neither depends on a
particular tree shape nor gets absorbed by a larger one. Growing the budget raises
$\tau$ for both drafters, but past $B=48$ the added verification cost outweighs the
added acceptance, so speedup peaks at our default for Parallel Reconsideration
($2.29\times$) and saturates for EAGLE-3.1 ($2.18\times$ to $2.20\times$). Widening $B_w$
beyond $6$ leaves $\tau$ unchanged for both and only costs speedup; narrowing it to $4$
costs speedup most for Parallel Reconsideration, whose speedup margin nearly vanishes
there ($2.112\times$ against $2.105\times$). Offer width behaves similarly: $k=4$
and $k=8$ keep the acceptance gain but lose speedup relative to $k=6$. The default is thus the best configuration for
Parallel Reconsideration, and it beats EAGLE-3.1 at every configuration tried,
including the baseline's own best ($2.29\times$ against $2.20\times$ at $B=64$).

\subsection{Reconsideration on top of a different drafter}
\label{app:exp_hass}

Reconsideration adds a second head and gate to a one-pass drafter, so it is not tied to
EAGLE-3.1. We test it on top of another speculative decoding method,
HASS~\citep{zhang2025hass}. We train
HASS and HASS with reconsideration on Llama-3.1-8B, compare them under the tree configuration of Table~\ref{tab:main_results} ($B=48$,
$k=6$, $B_w=6$, $\gamma=8$), and evaluate on all seven benchmarks at temperature $0$
(Table~\ref{tab:hass}).

\begin{table*}[t]
    \centering
    \caption{\textbf{Parallel Reconsideration on top of HASS.} Acceptance length $\tau$
        and speedup over autoregressive decoding on Llama-3.1-8B at temperature $0$,
        both drafters taken at the same training checkpoint and served
        under the tree configuration of Table~\ref{tab:main_results} ($B=48$).
        $\Delta$ is the relative gain of adding reconsideration to HASS.
    }
    \label{tab:hass}
    \resizebox{\linewidth}{!}{
        \begin{tabular}{lcccccccccccccccc}
            \toprule
                               & \multicolumn{6}{c}{\textbf{Math}} & \multicolumn{6}{c}{\textbf{Code}} & \multicolumn{2}{c}{\textbf{Chat}} &                                                                                                                                                                                                                                                                                                                           \\
            \cmidrule(lr){2-7}\cmidrule(lr){8-13}\cmidrule(lr){14-15}
                               & \multicolumn{2}{c}{GSM8K}         & \multicolumn{2}{c}{MATH-500}      & \multicolumn{2}{c}{AIME25}        & \multicolumn{2}{c}{HumanEval} & \multicolumn{2}{c}{MBPP} & \multicolumn{2}{c}{LCB} & \multicolumn{2}{c}{MT-Bench} & \multicolumn{2}{c}{\textbf{Avg.}}                                                                                                                                                                     \\
            \cmidrule(lr){2-3}\cmidrule(lr){4-5}\cmidrule(lr){6-7}\cmidrule(lr){8-9}\cmidrule(lr){10-11}\cmidrule(lr){12-13}\cmidrule(lr){14-15}\cmidrule(lr){16-17}
            \textbf{Method}    & $\tau$                            & Speedup                           & $\tau$                            & Speedup                       & $\tau$                   & Speedup                 & $\tau$                       & Speedup                           & $\tau$        & Speedup               & $\tau$        & Speedup               & $\tau$        & Speedup               & $\tau$            & Speedup               \\
            \midrule
            HASS               & 4.81                              & 2.03$\times$                      & 4.84                              & 2.11$\times$                  & 5.21                     & 2.27$\times$            & 5.59                         & 2.38$\times$                      & 5.34          & 2.38$\times$          & 4.68          & 1.90$\times$          & 4.64          & 1.97$\times$          & 5.02              & 2.15$\times$          \\
            \textbf{HASS + PR} & \textbf{5.17}                     & \textbf{2.12$\times$}             & \textbf{5.21}                     & \textbf{2.21$\times$}         & \textbf{5.88}            & \textbf{2.49$\times$}   & \textbf{5.86}                & \textbf{2.45$\times$}             & \textbf{5.67} & \textbf{2.44$\times$} & \textbf{4.99} & \textbf{1.98$\times$} & \textbf{4.97} & \textbf{2.02$\times$} & \textbf{5.39}     & \textbf{2.24$\times$} \\
            \quad $\Delta$     & $+7.6\%$                          & $+4.6\%$                          & $+7.7\%$                          & $+4.6\%$                      & $+12.7\%$                & $+9.6\%$                & $+4.7\%$                     & $+2.9\%$                          & $+6.0\%$      & $+2.4\%$              & $+6.6\%$      & $+3.9\%$              & $+7.2\%$      & $+2.6\%$              & $\mathbf{+7.5\%}$ & $\mathbf{+4.4\%}$     \\
            \bottomrule
        \end{tabular}
    }
\end{table*}

Parallel Reconsideration lifts HASS on every benchmark, by $\mathbf{7.5\%}$ in average acceptance
length ($5.02 \to 5.39$) and $\mathbf{4.4\%}$ in speedup ($2.15\times \to 2.24\times$),
comparable to its gain over EAGLE-3.1 on the same model
(Table~\ref{tab:main_results}). The gain is therefore not specific to one base drafter.
The largest gain is on AIME25 ($+12.7\%$ acceptance, $+9.6\%$ speedup).

\subsection{Which failures does reconsideration recover?}
\label{app:exp_recovery}

Figure~\ref{fig:diagnosis} is correlational: a decisive target does not by itself show
that a one-layer drafter could match it with more computation. As a partial check on the
reading of Section~\ref{sec:moti_insight}, we measure \emph{which} of the drafter's
failures a trained second pass recovers. Taking every token the one-pass drafter
rejects ($30.2\%$ of $3.1$M scored positions on the held-out split) and bucketing them
by the same $c^{\mathrm{d}}_p \times c^{\mathrm{t}}_p$ bands, we report the fraction of
each bucket reconsideration, trained on the same model, turns from wrong to right (Table~\ref{tab:recovery}).

\begin{table}[t]
    \centering
    \caption{\textbf{What reconsideration recovers, by bucket.}
        \emph{Top:} percentage of each bucket's rejected tokens that Parallel
        Reconsideration gets right, with the bucket's share of the reject pool in
        parentheses. Buckets are the $c^{\mathrm{d}}_p \times c^{\mathrm{t}}_p$ bands of
        Section~\ref{sec:moti_diagnosis}. \emph{Bottom:} mean change in top-$1$
        probability against the position's own one-pass distribution, on the tokens
        reconsideration recovers and on those it does not.}
    \label{tab:recovery}
    \resizebox{\linewidth}{!}{
        \begin{tabular}{lcccc}
            \toprule
                                                & \multicolumn{4}{c}{\textbf{target confidence} $c^{\mathrm{t}}_p$}                                                                        \\
            \cmidrule(lr){2-5}
            \textbf{drafter} $c^{\mathrm{d}}_p$ & $<0.5$                                                            & $0.5$--$0.9$    & $\ge 0.9$                & \textbf{all}            \\
            \midrule
            \multicolumn{5}{l}{\emph{recovered, \% of bucket (bucket's share of pool)}}                                                                                                    \\
            $<0.5$                              & 17.7\% (18.5\%)                                                   & 29.5\% (26.6\%) & \textbf{39.7\%} (17.0\%) & 28.8\% (62.1\%)         \\
            $0.5$--$0.9$                        & 11.3\% (6.7\%)                                                    & 23.4\% (13.6\%) & \textbf{42.0\%} (15.1\%) & 29.0\% (35.4\%)         \\
            $\ge 0.9$                           & 3.7\% (0.4\%)                                                     & 6.9\% (1.3\%)   & 27.1\% (0.8\%)           & 12.9\% (2.5\%)          \\
            \textbf{all}                        & 16.4\% (25.6\%)                                                   & 26.7\% (41.5\%) & \textbf{39.9\%} (32.9\%) & \textbf{28.5\%} (100\%) \\
            \midrule
            \multicolumn{5}{l}{\emph{$\Delta$ top-$1$ probability vs.\ own one-pass distribution}}                                                                                         \\
            recovered                           & $+0.008$                                                          & $+0.024$        & $\mathbf{+0.061}$        & $\mathbf{+0.039}$       \\
            not recovered                       & $+0.000$                                                          & $+0.002$        & $+0.004$                 & $+0.001$                \\
            \bottomrule
        \end{tabular}
    }
\end{table}

Reconsideration recovers \textbf{28.5\%} of all rejected tokens, and where it recovers
them supports the hypothesis. Recovery rises monotonically along the target axis
($16.4 \to 26.7 \to \textbf{39.9\%}$): it is governed by whether the target had a
decisive answer, not by how unsure the drafter sounded. The premature commitments of
Observation 2, drafted with $c^{\mathrm{d}}_p < 0.5$ where the target was decisive, are
recovered \textbf{39.7\%} of the time, while the both-uncertain tokens that Observation 2
identified as largely unrecoverable move least ($17.7\%$). A mechanism that merely
sharpened the existing distribution would help most where the drafter was already close; reconsideration instead spends itself on the
determined-but-missed tokens, which is the band Section~\ref{sec:moti_insight} set out to
reach.

The added confidence is selective in the same way (Table~\ref{tab:recovery}, bottom).
On the tokens it recovers, reconsideration raises the top-$1$ probability by
$\mathbf{+0.039}$, rising to $\mathbf{+0.061}$ where the target is decisive; on the
tokens in the same pool it fails to recover, the same measurement gives $+0.001$.
Each token is measured against its own one-pass distribution, which keeps
$c^{\mathrm{d}}_p$, the bucketing variable, out of the difference. This suggests the deeper pass
does not simply sharpen indiscriminately.

\section{Details on Related Work}
\label{app:related_work}

\paragraph{Speculative decoding.}
Speculative decoding accelerates autoregressive generation by drafting several tokens
cheaply and verifying them in a single target forward pass, accepting the longest prefix
the target agrees with so the output distribution is preserved exactly
\citep{leviathan2023fast, xia2023speculative}. Because the acceptance rule preserves
that distribution by construction, subsequent work improves speed along two axes: the
\emph{acceptance length}, how many drafted tokens survive verification per round, and the
\emph{drafting cost}, how much latency the draft itself adds. Our cost model
(Section~\ref{sec:motivation}) is the standard amortized form of this trade-off, and
Parallel Reconsideration targets the first axis without paying on the second.

\paragraph{Feature-level drafting.}
The original formulation pairs the target with a smaller standalone language model as the
drafter, which must be trained or found separately for each target. Self-drafting instead
derives the draft from the target's own computation. The EAGLE family drafts at the
feature level, autoregressing on a combination of hidden features from several target
layers with a single lightweight decoder layer and verifying candidates with dynamic
draft trees
\citep{li2024eagle, li2024eagle2, li2025eagle3}, most recently corrected for the
attention drift that degrades the drafter as the speculation chain
lengthens~\citep{eldenk2026attention}; a related line attaches lightweight
prediction heads directly to the target's last hidden state, either independent (Medusa
heads) \citep{cai2024medusa} or made sequentially dependent to raise acceptance (Hydra
heads) \citep{ankner2024hydra}. Parallel Reconsideration is built on the EAGLE-style
feature-level drafter but is orthogonal to this design axis: it adds a second guess at
uncertain positions on top of the existing one-pass draft, and applies to any drafter
that exposes a reusable draft feature.

\paragraph{Adaptive drafter compute and draft budgets.}
Several lines of work already vary how much the drafter computes per token.
Self-speculative decoding drafts with a subset of the target's own layers: Draft \&
Verify skips intermediate layers during drafting \citep{zhang2024draftverify}, LayerSkip
trains the target for early exit and drafts from its shallow layers
\citep{elhoushi2024layerskip}, and Kangaroo drafts from a shallow sub-network and exits
drafting early once confidence drops \citep{liu2024kangaroo}. These methods adapt the
\emph{depth} of a sequential draft, so extra compute on a token is extra sequential
latency; Parallel Reconsideration keeps the one-layer feature-level drafter and runs the
extra pass beside the ordinary step. ReDrafter replaces the draft heads with a recurrent
drafter conditioned on the target's hidden state \citep{cheng2024redrafter}, which still
uses one recurrent step per drafted token. Sequoia and SpecExec allocate a fixed
draft-tree budget unevenly across positions, placing more candidates where they are most
likely to be accepted \citep{chen2024sequoia, svirschevski2024specexec}, as EAGLE-2's
dynamic trees do with drafter confidence \citep{li2024eagle2}. These methods vary how many
candidates each position receives, not how each candidate is computed; reconsidered
candidates enter the same tree and so compose with them. A direct empirical comparison
with these methods is left to future work.

\paragraph{Training-free drafting.}
A complementary direction removes the trained drafter entirely. Lookahead decoding
generates and verifies $n$-gram continuations in parallel through Jacobi iteration,
requiring no auxiliary model or datastore \citep{fu2024lookahead}, and FAFO extends this
draftless $n$-gram candidate-pool paradigm by letting a compressed KV cache produce
lossy guesses that the same forward pass verifies \citep{le2026fafo}.
Such methods trade draft quality for zero training overhead; Parallel Reconsideration
takes the opposite stance, spending a small amount of otherwise-idle draft compute to
make each speculated position more accurate.

\paragraph{Parallel drafting.}
A drafter that autoregresses over $\gamma$ positions is itself sequential, so a line of
work replaces it with one that emits an entire block in a single forward pass. PARD
trains a target-independent draft model to predict several future tokens at once
\citep{an2025pard}, and diffusion-based drafters push this further: DFlash conditions a
block diffusion model on target features to produce a draft block in one pass and holds
the strongest reported speedups on this benchmark suite \citep{chen2026dflash}, while
DSpark couples a parallel backbone with a lightweight
sequential module to restore intra-block dependencies \citep{cheng2026dspark}. These
methods reduce the \emph{cost} axis by removing the drafter's sequential steps, whereas
Parallel Reconsideration raises the \emph{acceptance} axis by improving each drafted
position; a block-parallel drafter still commits to every position from a single pass, so
reconsideration composes with it.

\paragraph{Draft training objectives.}
Orthogonal to the drafter's architecture is what it is trained to optimize. Standard
drafters minimize per-token likelihood against the target, which is misaligned with the
sequence-level acceptance that verification actually rewards. Variational Speculative
Decoding recasts draft training as variational inference over draft paths, maximizing the
marginal probability of target acceptance rather than token likelihood
\citep{zou2026vsd}. Our training (Section~\ref{sec:training}) shares this acceptance-first
view but pursues it through a different mechanism: rather than reweighting a single draft
distribution, we train a second reconsideration pass to recover the specific positions
the one-pass draft gets wrong.

\paragraph{Looped and adaptive-compute transformers.}
The idea of spending extra passes to sharpen a prediction has a long history. Looped
transformers reuse a fixed block to emulate iterative computation
\citep{giannou2023looped}, universal transformers add a recurrent depth with a per-token
halting mechanism \citep{dehghani2019universal}, and adaptive-compute methods such as
PonderNet and CALM learn how many steps or layers each token needs
\citep{banino2021pondernet, schuster2022calm}. A recent looped language model, Ouro,
applies this recurrence to the target model during pre-training to improve its own
generation quality \citep{zhu2025ouro}. Our use of recurrence differs in purpose: we loop
not to make an autoregressive model reason better, but to repair the specific failure of
autoregressive \emph{drafting}, where the drafter must commit to a token before it has
seen enough context to place it, and we run the extra pass in parallel so it costs no
additional sequential step.

\section{Implementation details}
\label{app:impl}

\paragraph{Training.} Both drafters follow EAGLE-3's released recipe on UltraChat
\citep{ding2023ultrachat}, with no hyperparameter retuning for reconsideration. We use
AdamW ($\beta_1=0.9$, $\beta_2=0.999$, no weight decay) at a peak learning rate of
$10^{-4}$, cosine-annealed to zero after linear warmup over the first $1.5\%$ of steps,
with gradients clipped to norm $0.5$. Training runs in BF16 with FP32 master weights
and gradient checkpointing, sequences capped at $4{,}096$ tokens, per-device batch size
$1$ at data-parallel degree $4$ and no accumulation, giving an effective batch of four
sequences (roughly $16$k tokens) per step. The balance coefficient of
Eq.~\eqref{eq:fullloss} is $\beta=0.01$.

\paragraph{Benchmarks.} We evaluate each benchmark on its full set.

\begin{table}[h]
    \centering
    \caption{\textbf{Evaluation benchmarks.} Each is evaluated on its full set.}
    \label{tab:benchmarks}
    \begin{tabular}{llr l}
        \toprule
        \textbf{Category}     & \textbf{Benchmark} & \textbf{Prompts} & \textbf{Source}                               \\
        \midrule
        Chat                  & MT-Bench           & 80               & FastChat \texttt{mt\_bench}                   \\
        \midrule
        \multirow{3}{*}{Math} & GSM8K              & 1{,}319          & \texttt{openai/gsm8k} (main)                  \\
                              & MATH-500           & 500              & \texttt{HuggingFaceH4/MATH-500}               \\
                              & AIME25             & 30               & \texttt{math-ai/aime25}                       \\
        \midrule
        \multirow{3}{*}{Code} & HumanEval          & 164              & \texttt{openai/openai\_humaneval}             \\
                              & MBPP               & 500              & \texttt{google-research-datasets/mbpp} (full) \\
                              & LiveCodeBench      & 400              & \texttt{livecodebench/code\_generation}       \\
        \midrule
        \textbf{Total}        &                    & \textbf{2{,}993} &                                               \\
        \bottomrule
    \end{tabular}
\end{table}

\section{Limitations}
\label{app:limitations}

Reconsideration, in principle, composes with any drafter that exposes a
reusable draft feature, including the block-parallel drafters that are currently state of the art (Appendix~\ref{app:related_work}). DFlash~\citep{chen2026dflash} is
the natural next target: it reduces the cost of drafting while reconsideration raises
the acceptance of each drafted position, so the two act on different terms of
Eq.~\eqref{eq:costmodel} and there is no obvious reason they would conflict. Unlike EAGLE-style drafters, however, DFlash does
not rely on a token embedding to guide its feature drafting, so there is no embedding
input for the self-query to take the place of (Eq.~\eqref{eq:coforward}); applying
reconsideration on top would require substantial redesign, which we leave for future
work. Doing so also requires retraining a DFlash drafter with the extra
head and gate, and its training recipe has not been released, so we cannot
reproduce the baseline or train reconsideration on top of it. We
therefore demonstrate on EAGLE-3.1, whose recipe is publicly released, and leave the block-parallel combination open.

\section{The cost of co-forwarding, measured}
\label{app:coforward}

Section~\ref{sec:method_parallel} argues that a reconsideration pass shares the batch
that advances the draft, costing $\epsilon$ rather than a full draft step. We check
that claim directly. For each cell we form the per-cycle time
$t_{\mathrm{cycle}} = \tau / \text{throughput}$, the wall-clock milliseconds per
verification step, and compare Parallel Reconsideration against EAGLE-3.1 run for run
at temperature $0$. Table~\ref{tab:coforward} reports the ratio alongside the serial
reference of Section~\ref{sec:method_parallel}: run in series, the $N$
reconsidered positions of a $\gamma$-step draft add $N$ sequential draft steps, so drafting
takes $(\gamma+N)\,t_{\mathrm{step}}$ instead of $\gamma\,t_{\mathrm{step}}$, a factor
of $(\gamma+N)/\gamma$. That factor inflates only the drafting phase, so the serial
version's whole-cycle cost would lie between $1$ and $(\gamma+N)/\gamma$ depending on
drafting's share of the cycle.

\begin{table}[h]
    \centering
    \caption{\textbf{Per-cycle cost of co-forwarding on Qwen3-8B.}
        Measured ratio of $t_{\mathrm{cycle}} = \tau/\text{throughput}$ (time per
        verification step), Parallel Reconsideration over EAGLE-3.1 run for run at
        temperature $0$; this is a whole-cycle ratio (drafting plus verification). With
        $\gamma=8$ draft steps and $N$ reconsidered positions per draft, serializing would
        stretch drafting by $(\gamma+N)/\gamma$ (Section~\ref{sec:method_parallel}), an
        upper bound on the serial whole-cycle ratio. Co-forwarding keeps the measured ratio
        near $1$.}
    \label{tab:coforward}
    \resizebox{\linewidth}{!}{
        \begin{tabular}{ccccccc}
            \toprule
            \textbf{$B$} & \textbf{$B_w$} & \textbf{$N$} & \textbf{EAGLE-3.1}   & \textbf{PR}          & \textbf{measured} & \textbf{serial drafting} \\
                         &                &              & $t_{\mathrm{cycle}}$ & $t_{\mathrm{cycle}}$ & whole-cycle ratio & $(\gamma{+}N)/\gamma$    \\
            \midrule
            48           & 6              & 3            & 26.03                & 26.60                & 1.022$\times$     & 1.375$\times$            \\
            24           & 6              & 3            & 25.55                & 25.94                & 1.015$\times$     & 1.375$\times$            \\
            \bottomrule
        \end{tabular}
    }
\end{table}

With $\gamma=8$ and $N=3$ reconsidered positions, serializing would lengthen drafting by $37.5\%$; co-forwarded,
the same reconsiderations lengthen the whole cycle by only $1.5$--$2.2\%$ (ratios
$1.015$ and $1.022$ at $B=24$ and $B=48$). The co-forwarded rows thus add roughly $2\%$
to total cycle time (drafting plus verification), consistent with the
memory-bandwidth argument of Section~\ref{sec:method_parallel}.

\section{Supervising both heads}
\label{app:train_ablation}
This section expands on the two degenerate training objectives of
Section~\ref{sec:training} and why each collapses serving quality.

\paragraph{Supervising only the reconsidered head.} Training only
$\mathrm{head}(f_p^{r})$ for $r\ge 1$ is the naive way to apply a looped transformer
\citep{giannou2023looped, dehghani2019universal} to drafting: run the layer to its final
depth and supervise the last head, exactly as one would optimize a recurrent model for
its own output. It fails here for a reason specific to speculative decoding. The one-pass
head $\mathrm{head}(f_p^{0})$ is never trained, so the one-pass draft is free to drift and
nothing holds it to ordinary-drafter quality. Because the served tree is grown from the
one-pass feature and the reconsidered candidates only extend prefixes the one-pass draft
already reaches, a degraded one-pass front leaves the reconsidered branches with short or
empty prefixes to build on. At serving, this can pull acceptance length below the
one-pass-only baseline even if the reconsidered head looks accurate in isolation under
teacher forcing. This is why adapting loop transformer idea to the draft is non-trivial:
unlike a looped language model, whose only output is the final pass, the draft must keep
\emph{every} intermediate pass serving-ready, since the tree is grown from the one-pass
feature and reconsideration only adds to it.

\paragraph{Supervising both heads symmetrically.} If instead both
$\mathrm{head}(f_p^{0})$ and $\mathrm{head}(f_p^{1})$ are trained to match the target with
equal weight, gradient descent has no incentive to make the two features differ, and it
is free to settle on the trivial optimum $f_p^{1}\!\approx\!f_p^{0}$. The reconsidered
token then echoes the one-pass token: the second pass is real compute that adds no new
candidate, and acceptance length can do no better than the one-pass baseline.

The recipe of Section~\ref{sec:training} avoids both failures by holding the one-pass
draft to full quality with an unconditional loss while supervising the reconsidered
feature only at the positions and depths where it is actually used, so $f_p^{1}$ is
pushed away from $f_p^{0}$ exactly where a second guess can help.

\section{Marginal vs.\ per-position entropy}
\label{app:entropy}

The balance term of Eq.~\eqref{eq:fullloss} is not the obvious entropy. The halt
distribution induces two: the per-position entropy averaged over positions,
$\mathbb{E}_p[H(q)]$, and the entropy of the marginal usage
$\bar q = \mathbb{E}_p[q]$. We reward the marginal $H(\bar q)$, and a clean
decomposition says why.

\begin{proposition}[Marginal vs.\ per-position entropy]
    \label{prop:entropy}
    For any distribution of per-position halt distributions $q$,
    \begin{align}
        H(\bar q) & = \underbrace{I(d;p)}_{\text{spread across depths}}
        \;+\; \underbrace{\mathbb{E}_p\big[H(q)\big]}_{\text{per-position indecision}}
        \nonumber                                                       \\[-2pt]
                  & \ge\; \mathbb{E}_p\big[H(q)\big],
        \label{eq:entropydecomp}
    \end{align}
    where $d\sim q$ is the halt depth and $I(d;p)\ge0$ is its mutual information with
    the position. Consequently $H(\bar q)$ depends on the routing only through the
    marginal $\bar q$: it is \emph{indifferent} between a deterministic routing that
    sends different positions to different depths ($I(d;p)$ large,
    $\mathbb{E}_p[H(q)]=0$) and an undecided one that splits every position across depths
    ($I(d;p)=0$, $\mathbb{E}_p[H(q)]$ large), so long as the two induce the same
    $\bar q$.
\end{proposition}

\begin{proof}[Proof of Proposition~\ref{prop:entropy}]
    Fix a batch of positions $p$, each with a
    halt distribution $q(\cdot\mid p)$ over depths $r \in \{0,1,\dots,R\}$, and let $p$
    be drawn uniformly over the batch. Define the joint distribution of position and
    halt depth $d$,
    \begin{equation}
        \Pr(p, d=r) = \Pr(p)\, q(r \mid p),
    \end{equation}
    and the marginal usage $\bar q(r) = \mathbb{E}_p[\,q(r\mid p)\,] = \Pr(d=r)$.

    \paragraph{Decomposition.} The entropy of the marginal is the entropy of $d$, which
    by the chain rule of mutual information decomposes as
    \begin{align}
        H(\bar q) = H(d)
         & = I(d;p) + H(d\mid p)                                  \nonumber \\
         & = I(d;p) + \mathbb{E}_p\big[H(q(\cdot\mid p))\big],
        \label{eq:app_decomp}
    \end{align}
    since $H(d\mid p) = \sum_p \Pr(p)\,H(q(\cdot\mid p)) = \mathbb{E}_p[H(q)]$. Because
    $I(d;p)\ge 0$, this gives $H(\bar q) \ge \mathbb{E}_p[H(q)]$, which is exactly the
    Jensen inequality $H(\mathbb{E}_p[q]) \ge \mathbb{E}_p[H(q)]$ applied to the concave
    entropy functional; \eqref{eq:app_decomp} identifies the gap as the mutual
    information $I(d;p)$. This proves \eqref{eq:entropydecomp}.

    \paragraph{Indifference.} Both terms in \eqref{eq:app_decomp} are nonnegative and
    $H(\bar q) \le \log(R{+}1)$, with equality iff $\bar q$ is uniform. Since
    $H(\bar q)$ is a function of $\bar q$ alone, it cannot distinguish two routings
    that induce the same marginal, however differently they allocate mass per position.
    Two extremes make this concrete. The \emph{deterministic} routings --- $q(\cdot\mid
        p)$ a point mass at some depth $d(p)$, with $d(p)$ distributed uniformly across the
    batch --- attain $H(\bar q)=\log(R{+}1)$ with $\mathbb{E}_p[H(q)] = 0$ and
    $I(d;p)=\log(R{+}1)$. The fully undecided routing --- every $q(\cdot\mid p)$ uniform
    --- attains the same $H(\bar q)=\log(R{+}1)$ with the decomposition reversed,
    $\mathbb{E}_p[H(q)]=\log(R{+}1)$ and $I(d;p)=0$. The balance term scores both
    identically, so it neither rewards nor penalizes per-position indecision; it penalizes
    only a marginal that has collapsed onto a subset of the depths. By contrast,
    regularizing $\mathbb{E}_p[H(q)]$ directly is maximized only by the second extreme,
    forcing per-position indecision.
\end{proof}

\paragraph{Practical note.} We estimate $\bar q$ per minibatch, so the balance term
rewards spread across the depths \emph{used within a batch} rather than a global
marginal; with the small coefficient $\beta=0.01$ this is enough to keep every depth
populated without measurably softening the per-position halt decisions.

\end{document}